\documentclass[1p,times]{elsarticle}

\usepackage{amssymb}
\usepackage{booktabs}
\usepackage{longtable}
\usepackage[table,xcdraw]{xcolor}
\usepackage{multirow}
\usepackage{subcaption}
\usepackage{pdflscape}
\usepackage{array,longtable}
\usepackage{multicol}
\usepackage{hyperref}
\usepackage{parskip}
\usepackage{amsmath}
\usepackage{graphicx}
\usepackage{fancyhdr}
\usepackage{vmargin}
\usepackage{mwe}
\usepackage{layout}
\usepackage{verbatim}
\usepackage{caption}
\usepackage{soul}
\usepackage{smartdiagram}
\usepackage{threeparttable}
\usepackage{threeparttablex}
\usepackage[final]{pdfpages}
\usepackage{natbib}
\usepackage{float}
\usepackage{scalerel}
\usepackage{booktabs} 
\usepackage{tabularx}
\usepackage{booktabs}
\usepackage{rotating}
\usepackage{graphicx}
\usepackage{siunitx}    

\journal{Elsevier}

\biboptions{authoryear}

\begin{document}

\setlength\parindent{0pt}

\begin{frontmatter}

\title{Adversarial Training for Tabular Credit Scoring: A Multi-Attack Robustness Evaluation in P2P Lending}

\author[inst1]{Gijs A. F. Niewzwaag}
\ead{g.a.f.niewzwaag@student.utwente.nl}
\author[inst1]{Marijn G. S. Veth}
\ead{m.g.s.veth@student.utwente.nl}
\author[inst1]{Manuele Massei}
\ead{manuele.massei@utwente.nl}
\author[inst1]{Marcos R. Machado\corref{mycorrespondingauthor}}
\cortext[mycorrespondingauthor]{Corresponding author: Marcos R. Machado (m.r.machado@utwente.nl, Tel. +31 534899045).}
\ead{m.r.machado@utwente.nl}

\affiliation[inst1]{organization={University of Twente},
addressline={Faculty of Behavioural, Management and Social Sciences, Department of High-Tech Business and Entrepreneurship},
city={AE Enschede},
postcode={7500},
country={Netherlands}}

\begin{abstract}
Machine learning-based credit scoring is increasingly central to Peer-to-Peer (P2P) lending, yet its resilience to adversarial manipulation—where applicants strategically alter self-reported inputs to secure favourable decisions—remains poorly understood. Most adversarial-robustness evidence comes from image and text domains and evaluates a single attack against a matching defence, offering little guidance on how defences generalise across attack types in tabular credit data. We address this with a systematic train–test robustness benchmark on a large Lending Club subset, spanning three model families (logistic regression, a feed-forward neural network, and a transformer for tabular data) and four attacks confined to applicant-mutable features: Fast Gradient Sign Method (FGSM), Projected Gradient Descent (PGD), Salt-and-Pepper (S\&P) noise, and DeepFool plus a mixed-attack regime. Across a full grid evaluated with stratified cross-validation, adversarial training sharply improves robustness against the attack it is trained on and transfers well within the gradient-based family, but transfers weakly to non-gradient corruption—so single-attack defences overstate real-world resilience. Mixed training delivers the most balanced robustness across heterogeneous attacks while preserving clean-test performance, supporting multi-attack stress testing in credit-model governance.
\end{abstract}


\end{frontmatter}


\section{Introduction}
\label{sec:intro}

Access to affordable credit is essential for economic mobility and social inclusion. Yet, globally, more than 1.4 billion adults remain unbanked and lack access to formal financial services \citep{WorldBank2025}. Even in advanced economies, gaps persist, particularly among low-income households, immigrants, and minority groups. Roughly 11\% of adults in the United States are either credit invisible or unscored, with Black and Hispanic communities disproportionately affected \citep{Fed2023}. Digital lending channels and alternative finance have expanded credit provision, but they also intensify reliance on automated decision systems whose reliability depends on data integrity and model robustness.

Credit risk assessment remains a cornerstone of financial decision-making for both traditional institutions and emerging actors in alternative finance \citep{djeundje2021enhancing}. While commercial banks have historically relied on proprietary scoring models and regulated data sources, Peer-to-Peer (P2P) lenders, crowdfunding platforms, and fintech startups increasingly leverage alternative data and Machine Learning (ML) models to evaluate creditworthiness \citep{zhang2020credit, emekter2015evaluating}. Irrespective of the institutional setting, the core task is to infer the likelihood that a borrower will repay a loan based on available information \citep{boyes1989econometric}. In practice, these inferences can be highly sensitive to feature engineering choices, model class, and noise or errors in the input data, making robustness a central concern for any operational credit scoring pipeline \citep{emekter2015evaluating}.

A growing and under-addressed dimension of robustness in credit scoring is vulnerability to deliberate input manipulation. As ML-based scoring becomes embedded in digital application flows, the attack surface expands: adversaries may strategically perturb input variables to obtain more favourable decisions or to degrade a lender's risk models \citep{straus2025explaining, schwab2025mitigating}. Such perturbations do not need to be large to be consequential; they can be small, targeted, and designed to exploit the geometry of a model's decision boundary \citep{straus2025explaining, schwab2025mitigating}. This connects credit scoring directly to Adversarial Machine Learning (AML), a field that studies how models behave under adversarially crafted inputs and how defences such as adversarial training can mitigate these risks \citep{straus2025explaining, schwab2025mitigating}.

The rapid adoption of AI in credit underwriting has also prompted regulatory and supervisory attention. Credit scoring is widely considered a high-stakes application because it can constrain individuals' financial opportunities and may lead to systemic risks when deployed at scale. In the European Union, the AI Act\footnote{https://digital-strategy.ec.europa.eu/en/policies/regulatory-framework-ai} designates credit scoring as a high-risk AI application, introducing obligations for risk management, documentation, transparency, and human oversight throughout the model lifecycle. Complementary legal regimes, including GDPR\footnote{https://gdpr-info.eu/} provisions on automated decision-making and long-standing anti-discrimination and consumer protection rules, further shape how scoring models can be developed, monitored, and audited. Other jurisdictions, including the United States\footnote{https://www.congress.gov/crs-product/R48555}, Brazil\footnote{https://www.gov.br/mcti/pt-br/acompanhe-o-mcti/transformacaodigital/arquivosinteligenciaartificial/}, and China\footnote{https://www.twobirds.com/en/capabilities/artificial-intelligence/ai-legal-services/ai-regulatory-horizon-tracker/china}, have likewise advanced governance frameworks for AI-enabled decision-making in finance. These developments underscore that credit scoring pipelines must be not only accurate under benign conditions, but also resilient to input perturbations and auditable under realistic threat scenarios.

From a methodological perspective, AML offers a diverse toolbox of attacks and defences. Gradient-based attacks such as the Fast Gradient Sign Method (FGSM) and Projected Gradient Descent (PGD) craft perturbations by exploiting local loss gradients, often under explicit norm constraints \citep{naseem2024trans, naqvi2023adversarial, gupta2018cnn}. Boundary-seeking methods such as DeepFool aim to identify small, targeted changes that flip predictions \citep{moosavi2016deepfool}, while non-gradient perturbations (e.g., noise-based manipulations such as Salt-and-Pepper) can approximate data corruption or opportunistic manipulation without requiring gradient access \citep{azzeh2018salt}. On the defence side, adversarial training is a prominent approach, but its effectiveness depends on the attack model used during training and the extent to which robustness transfers to unseen attacks \citep{naseem2024trans, naqvi2023adversarial, gupta2018cnn, moosavi2016deepfool, azzeh2018salt}. In applied financial settings, evidence remains limited on exhaustive, hybrid evaluations that mix multiple attacks during training and systematically assess cross-attack generalisation in out-of-sample data. This gap is consequential because operational attackers are unlikely to adhere to a single threat model, and robustness evaluated under one-to-one attack--defence pairings may overstate real-world resilience.

This study addresses these challenges by investigating whether adversarial training can enhance the robustness of credit scoring models in P2P lending. Using the Lending Club dataset\footnote{https://www.lendingclub.com/} and state-of-the-art AML techniques, we systematically evaluate model performance under multiple adversarial attack families, including gradient-based and non-gradient perturbations. The analysis is guided by the following central research question: \textit{To what extent can adversarial training enhance the robustness of credit scoring models in P2P lending against both gradient-based and non-gradient adversarial attacks?}.

The main contributions of this paper are threefold. First, we provide a comprehensive empirical evaluation of AML in P2P credit risk assessment by testing multiple attack algorithms, DeepFool, Salt-and-Pepper noise, FGSM, and PGD, thereby covering qualitatively different perturbation mechanisms, and by constraining perturbations to applicant-mutable features to reflect realistic manipulation of self-reported inputs. Second, we introduce and assess mixed training--testing strategies that intentionally vary the type of attack used during adversarial training and the type encountered at evaluation time. This design enables a systematic analysis of cross-attack generalisation (i.e., whether training for one threat improves robustness to other threats) and whether robustness transfers across perturbation families. Third, by benchmarking adversarially trained models against standard baselines on real-world lending data, we derive practical guidance on which modelling and training choices preserve predictive performance when inputs are strategically manipulated.

The remainder of the paper is structured as follows. Section \ref{sec:litreview} reviews related literature on credit scoring, ML models, and adversarial machine learning. Section \ref{sec:materials} outlines the methodological approach, including the CRISP-ML process, adversarial techniques, dataset, preprocessing, and model implementation. Section \ref{sec:results} presents and discusses the empirical findings. Finally, Section \ref{sec:conclusion} concludes with key insights and directions for future research.

\section{Literature Review}
\label{sec:litreview}

This section positions the study within three interrelated directions: (i) credit risk modelling in traditional and ML settings, with an emphasis on P2P lending; (ii) AML and robustness for tabular decision systems; and (iii) the security, ethical, and regulatory constraints that shape the deployment of AI in credit underwriting. We conclude by synthesising the key research gaps and the how this study adds to the literature in this field.

\subsection{Traditional versus ML Methods for Credit Risk Assessment}

Credit risk assessment has traditionally relied on structured, rule-based methodologies developed and refined over decades \citep{boyes1989econometric, masters1982rasch, altman1980commercial}. Classical approaches, often based on logistic regression or expert-driven scorecards, primarily use curated data from credit bureaus, loan performance histories, and manually engineered features \citep{Siddiqi2017}. Their strengths lie in interpretability, stability, and regulatory familiarity. However, such models are limited in their ability to adapt to complex borrower behaviours or capture subtle patterns in diverse and dynamic financial environments.

By contrast, ML approaches offer greater flexibility and predictive capacity. ML models can identify nonlinear relationships and high-dimensional feature interactions, thereby improving decision-making speed and accuracy \citep{MachadoKarray2022_hybrid, GoldmannMachado2025_hybrid}. Moreover, they can integrate alternative data sources such as mobile usage, utility payments, and digital footprints, which are especially valuable for applicants lacking formal credit histories \citep{Svitla2024}. This data-driven adaptability has positioned ML as a powerful tool for expanding credit access in emerging and digital-first economies \citep{van2025can}.

Key distinctions between traditional and ML-based methods include \citep{MachadoKarray2022_hybrid, MachadoEtAl2025_analytical, GoldmannMachado2025_hybrid, boyes1989econometric, masters1982rasch, altman1980commercial, MokheleliMuseba2023}:

\begin{itemize}
  \item Data Sources: Traditional models rely on regulated and static datasets, while ML models expand the scope to dynamic, heterogeneous, and often real-time data streams.
  \item Automation: ML enables scalable and automated decision-making through APIs and intelligent scoring systems.
  \item Pattern Recognition: ML uncovers complex, nonlinear relationships often missed by linear models, improving performance across heterogeneous borrower pools.
  \item Bias Handling: Depending on design, ML can either amplify or mitigate biases. When fairness constraints are incorporated, ML has been shown to reduce discrimination more effectively than rule-based approaches.
\end{itemize}

Empirical evidence consistently shows that ML models, particularly ensemble methods such as random forests and gradient boosting, outperform traditional techniques across metrics including precision, recall, AUCPR, and F1-score \citep{MokheleliMuseba2023, west2000neural}. Yet, these improvements come with trade-offs in interpretability, validation complexity, and regulatory compliance. Since credit decisions carry significant societal and legal implications, robustness and explainability remain critical dimensions alongside accuracy \citep{HengSubramanian2022, PatilIyerEtAl2024, NallakaruppanChaturvediEtAl2024}. These tensions motivate closer attention to robustness in adversarial settings, particularly in digital lending contexts where inputs may be noisy or strategically manipulated \citep{yan2015signaling, caldieraro2018strategic}.

\subsubsection{Credit Risk Assessment in P2P Lending}

P2P lending platforms have redefined credit intermediation by directly connecting borrowers with lenders \citep{liang2020analyzing, zhang2020credit, laia2025role, del2025exploring}. These platforms rely heavily on algorithmic risk assessment to replace or supplement traditional underwriting. Compared to banks, which are subject to conservative regulation and rely on standardised scoring, P2P platforms often employ greater flexibility in data sourcing, feature engineering, and model experimentation. This allows rapid innovation, but also heightens exposure to risks concerning model quality, fairness, and security \citep{ZhangYu2024}. In particular, the predominantly digital and scalable nature of P2P underwriting can increase the likelihood that application inputs are incomplete, inconsistent, or intentionally distorted \citep{chen2025too, del2025exploring}.

\citet{ZhangYu2024} propose a framework for understanding P2P credit risk models along three dimensions: classification algorithms, data traits, and learning methods. Among these, classification algorithms remain central. As summarised in Table~\ref{tab:classification-algos}, based on \citet{ZhangYu2024}'s study, there are four broad algorithm categories that dominate the literature on this topic.

\begin{table}[htb]
  \fontsize{10}{10}\selectfont  
  \centering
  \caption{Summary of classification algorithms.}
  \label{tab:classification-algos}
  \begin{tabularx}{\textwidth}{@{}
      >{\bfseries}l
      X
      X
    @{}}
    \toprule
    Classification Algorithms
      & Advantages
      & Disadvantages \\
    \midrule
    Traditional Single Classifiers
      & Interpretable, low complexity
      & Limited capacity for nonlinear data \\
    Intelligent Single Classifiers
      & High accuracy on structured inputs
      & Sensitive to noise and parameter tuning \\
    Hybrid Multiple Classifiers
      & Better generalization
      & High training complexity, reduced interpretability \\
    Ensemble Multiple Classifiers
      & Robust and scalable
      & Black-box nature, difficult to audit \\
    \bottomrule
  \end{tabularx}
\end{table}

Beyond model choice, feature engineering is pivotal. In P2P contexts, borrowers often lack extensive credit histories, requiring reliance on diverse predictors. A review by \citet{SegunYemiPetersAdewumi2024} identifies five recurring categories: (i) demographic variables, (ii) financial history, (iii) employment and income, (iv) loan and application characteristics, and (v) macroeconomic context. The richness of these features highlights both the opportunities and vulnerabilities of P2P models. Models that leverage diverse data may achieve high predictive power, but they can also become more susceptible to perturbations and gaming, especially when features are self-reported or weakly verifiable \citep{diao2025misreporting, wang2025effectiveness}, a key motivation for this paper’s research question.

In this study, a Neural Network (NN) is selected as the baseline model, reflecting its ability to handle structured, high-dimensional inputs and its compatibility with adversarial training \citep{west2000neural, HeOuyangWenLiuMoreira2025}. This methodological choice directly supports our objective of assessing how adversarial robustness can be integrated into practical P2P credit risk modelling.

\subsection{Adversarial ML and Robustness}

Adversarial Machine Learning (AML) investigates the vulnerability of ML systems to strategically crafted perturbations designed to mislead predictions \citep{sarkar2018, hariprasad2025}. Such attacks are typically subtle and may remain undetected by standard validation processes that assume i.i.d.\ test conditions \citep{MadryEtAl2017, naqvi2023adversarial}. Given the growing deployment of ML in high-stakes domains such as finance and healthcare, understanding and mitigating adversarial risk has become a critical research frontier \citep{JedrzejewskiEtAl2024, PelekisEtAl2025}.

Attacks are commonly classified along four dimensions \citep{JedrzejewskiEtAl2024, hariprasad2025}:

\begin{itemize}
  \item Attacker Knowledge: White-box (full access), black-box (no access), or gray-box (partial access).
  \item Specificity: Targeted (forcing a particular misclassification) or untargeted (any misclassification).
  \item Objective: Integrity (misleading predictions), availability (denying service), or privacy (extracting sensitive data).
  \item Phase: Evasion (during inference) or poisoning (during training).
\end{itemize}

This study implements five representative adversarial configurations: the Fast Gradient Sign Method (FGSM), Projected Gradient Descent (PGD), Salt-and-Pepper (S\&P) noise, and DeepFool as a boundary-seeking attack. In addition, we employ a mixed training scheme that combines these techniques to more rigorously stress-test model robustness  \citep{moosavi2016deepfool, naqvi2023adversarial, MadryEtAl2017}. FGSM and PGD represent white-box evasion attacks exploiting gradient information \citep{naqvi2023adversarial, MadryEtAl2017}, whereas S\&P simulates non-gradient corruption through high-variance perturbations \citep{azzeh2018salt}. Together, these attacks provide a testbed to evaluate the resilience of conventional and adversarially trained models in both gradient-based and non-gradient scenarios \citep{kabir2025giids, niu2025enhancing}.

Robustness in ML refers to the stability of predictions under unexpected, noisy, or adversarial inputs \citep{hariprasad2025}. In credit risk assessment, robustness failures may lead to serious consequences, such as granting loans to fraudulent applicants or mispricing risk under strategic behaviour \citep{hess2025machine, jakubik2025improving}. Thus, robustness must be considered alongside accuracy as a core performance dimension \citep{Dubrov2023}. Adversarial training, which augments datasets with perturbed examples, is among the most established defences. It encourages models to form smoother, more resilient decision boundaries \citep{PelekisEtAl2025, MadryEtAl2017}, though often at the cost of reduced clean-data accuracy.

By evaluating how NNs trained with adversarial examples perform relative to standard models, this paper directly addresses the central research question on the potential of adversarial training to enhance robustness in P2P credit risk assessment.

\subsection{Security, Ethics, and Regulatory Aspects of Credit Risk Assessment}

The deployment of AI in credit scoring is not only a technical challenge but also an ethical and regulatory one \citep{mathen2025toward}. Algorithmic decisions must safeguard fairness, privacy, and accountability to prevent exacerbating existing inequalities.

Adversarial attacks can compromise fairness by enabling certain applicants to exploit vulnerabilities, thereby disadvantaging honest borrowers \citep{MedingHagendorff2024}. Moreover, biases may be amplified if marginalised groups are disproportionately misclassified under attack \citep{KozodoiJacobLessmann2021}. Privacy risks arise when adversaries attempt to infer sensitive information or reconstruct training data, raising significant legal concerns under data protection regimes such as the GDPR. Beyond individual harms, robustness failures can undermine platform trust and increase losses through systematic misclassification of risky loans \citep{huang2025crystal, atahau2025p2p}.

Regulatory frameworks increasingly demand robust and transparent systems. In the EU, the AI Act designates credit scoring as ``high-risk,'' requiring strict documentation, explainability, and human oversight \citep{AIAct2024, EUParliament2025}. In the United States, laws such as the Equal Credit Opportunity Act (ECOA) and the Fair Credit Reporting Act (FCRA) impose accountability and anti-discrimination obligations \citep{FedFairLend2024}. Meeting these standards requires methods that strengthen robustness while preserving interpretability. Recent work suggests that integrating robustness evaluation into model governance and assurance processes can support compliance by making performance more stable and auditable under realistic stress scenarios \citep{huang2025crystal, atahau2025p2p, KozodoiJacobLessmann2021, sarkar2018}. Advances in defence-oriented methods (including explanation-guided defences in credit settings) further indicate that robustness and accountability need not be mutually exclusive \citep{straus2025explaining}.

\subsection{Research Gaps}
\label{subsec:research_gaps}

Despite growing interest in AML, the credit risk literature still provides limited evidence on how to systematically evaluate and build robustness for structured credit datasets—particularly in P2P lending contexts. Three gaps are especially salient.

First, many AML studies are developed around image or text domains, and their evaluation protocols do not readily transfer to structured tabular decisioning where feature constraints, discreteness, and plausibility requirements matter \citep{azzeh2018salt, naqvi2023adversarial}. While recent benchmarking efforts for tabular attacks improve comparability and tooling, they are typically domain-agnostic and do not focus on credit-specific constraints or regulatory expectations \citep{HeOuyangWenLiuMoreira2025}. This limits direct guidance for financial institutions and P2P platforms seeking to operationalise AML stress testing.

Second, work that does examine robustness in finance often evaluates a narrow set of attacks or relies on single-pair attack--defence assessments (e.g., training and testing on the same attack family), which provides limited insight into cross-attack generalisation \citep{MadryEtAl2017, PelekisEtAl2025, JedrzejewskiEtAl2024}. In deployment, however, attackers may adapt, combine perturbation styles, or exploit the absence of gradient access, making robustness claims under one threat model potentially brittle \citep{NallakaruppanChaturvediEtAl2024, sarkar2018}. A comprehensive assessment therefore requires testing multiple AML algorithms spanning different perturbation mechanisms and attacker assumptions.

Third, the literature rarely explores hybrid training--testing strategies that intentionally mix training attacks and evaluation attacks, or that assess whether robustness learned under one perturbation family transfers to others in out-of-sample data. This is a central practical question for model governance: if organisations can only afford to train against a subset of attacks, which subset (or specific model) offers the best overall resilience? Addressing this gap aligns directly with the contributions stated in Section~\ref{sec:intro}, where we propose an exhaustive multi-attack evaluation and mixed training--testing design.

To make these gaps explicit and to connect prior work to our contribution, Table~\ref{tab:gap_table} summarises representative studies and indicates whether they (i) address credit risk, (ii) focus on P2P settings, (iii) consider AML, (iv) evaluate multiple AML attacks, and (v) test mixed training--testing across different AML algorithms, the final row reports the characteristics of this study.

\begin{table}[htb]
\fontsize{10}{10}\selectfont 
\centering
\caption{Overview of representative related studies and coverage of AML robustness dimensions (illustrative mapping based on stated focus of each work).}
\label{tab:gap_table}
\begin{tabularx}{\textwidth}{@{}
l
>{\centering\arraybackslash}X
>{\centering\arraybackslash}X
>{\centering\arraybackslash}X
>{\centering\arraybackslash}X
>{\centering\arraybackslash}X
@{}}
\toprule
\textbf{Study} &
\textbf{Credit risk} &
\textbf{P2P} &
\textbf{AML} &
\textbf{Multiple attacks} &
\textbf{Mix train/test}\\
\midrule
\citet{boyes1989econometric} & Yes & No & No & No & No \\
\citet{Siddiqi2017}          & Yes & No & No & No & No \\
\citet{west2000neural}       & Yes & No & No & No & No \\
\citet{emekter2015evaluating}& Yes & Yes & No & No & No \\
\citet{zhang2020credit}      & Yes & Yes & No & No & No \\
\citet{liang2020analyzing}   & Yes & Yes & No & No & No \\
\citet{ZhangYu2024}          & Yes & No  & No & No & No \\
\citet{sarkar2018}           & Yes & Yes & Yes & Yes & No \\
\citet{Naidu2019}            & Yes & No  & Yes & No  & No \\
\citet{schwab2025mitigating} & Yes & Yes & Yes & No  & No \\
\citet{straus2025explaining} & Yes & Yes & Yes & No  & No \\
\citet{HeOuyangWenLiuMoreira2025} & No & No & Yes & Yes & No \\
\citet{JedrzejewskiEtAl2024}      & No & No & Yes & Yes & No \\
\citet{PelekisEtAl2025}           & No & No & Yes & Yes & No \\
\midrule
\textbf{This study} & \textbf{Yes} & \textbf{Yes} & \textbf{Yes} & \textbf{Yes} & \textbf{Yes} \\
\bottomrule
\end{tabularx}
\end{table}

Table \ref{tab:gap_table} highlights then that prior work establishes the value of ML for credit scoring and highlights the emerging relevance of AML. However, the literature provides limited guidance on exhaustive and hybrid robustness evaluation for P2P credit risk models. Our study addresses this gap by systematically testing multiple AML algorithms (DeepFool, S\&P, FGSM, PGD) and by evaluating mixed training--testing strategies to quantify cross-attack generalisation on real-world P2P lending data.

\section{Materials and Methods}
\label{sec:materials}

\subsection{Overview and Process Model}

\begin{figure}[htb]
  \centering
  \includegraphics[width=1\textwidth]{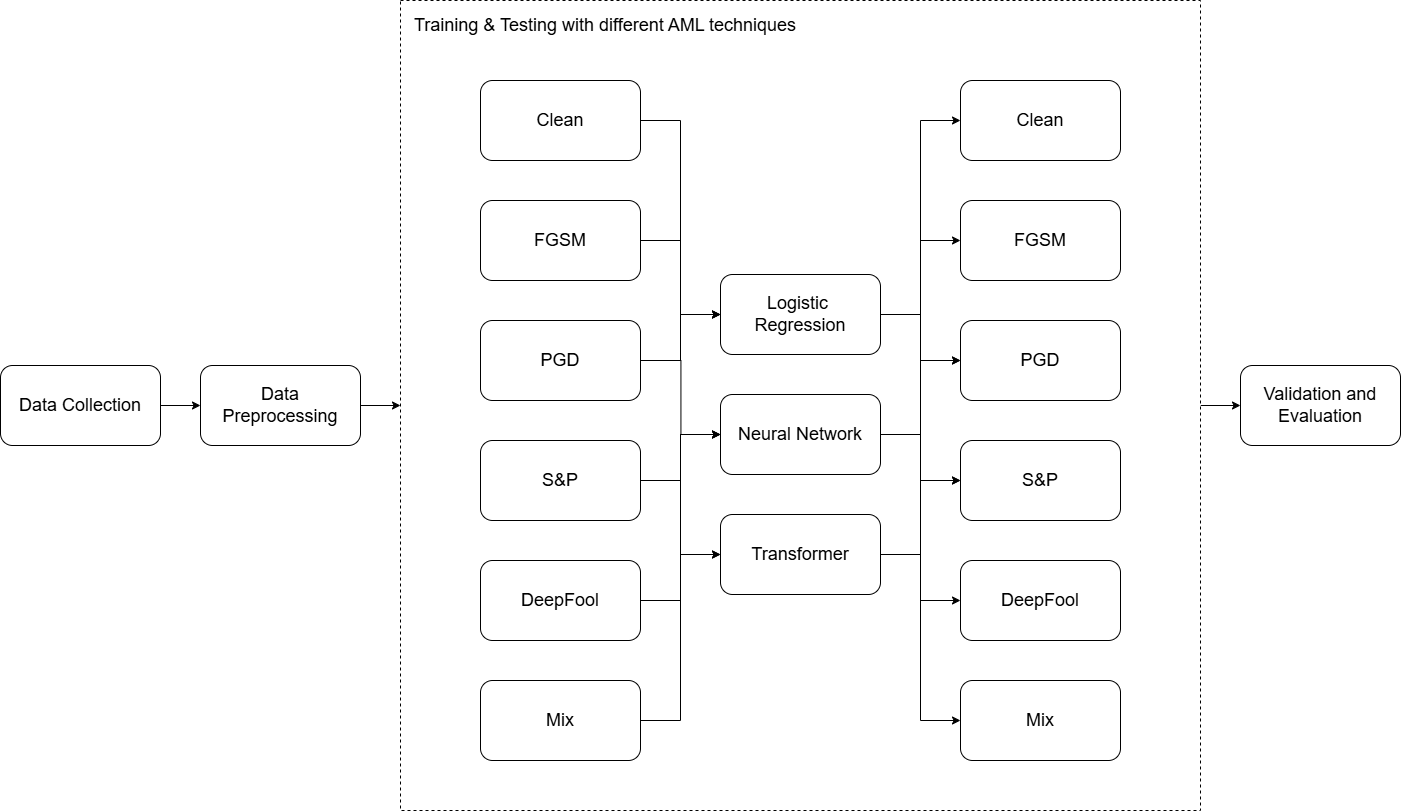}
  \caption{Experimental Setup of this Study.}
  \label{fig:flow chart set up}
\end{figure}

Our experimental pipeline is inspired by the CRISP--ML lifecycle: Business Understanding, Data Understanding, Data Preparation, Modelling, Evaluation, and (optional) Deployment, which extends CRISP-DM for predictive ML systems and emphasises quality assurance, including robustness checks \citep{studer2021towards, wirth2000crisp}. Rather than reproducing CRISP--ML verbatim, we translate its principles into a step-by-step experimental design tailored to adversarial robustness in P2P credit risk assessment. Figure~\ref{fig:flow chart set up} provides an overview, moving from Data Collection and Data Preprocessing (cleaning, encoding, feature engineering, outlier handling, and scaling) to modelling and evaluation.

We benchmark three model families representative of common approaches for tabular credit scoring \citep{gouvea2007credit, zhu2016predicting, machineenhancing}: a linear model (logistic regression), a neural model (feed-forward neural network), and a transformer-based model for tabular data. To emulate manipulation of applicant inputs at inference time, we generate perturbed samples using four adversarial techniques: Fast Gradient Sign Method (FGSM), Projected Gradient Descent (PGD), DeepFool, and Salt-and-Pepper (S\&P) noise \citep{azzeh2018salt, naseem2024trans, kabir2025giids}. FGSM, PGD, and DeepFool capture gradient-based (white-box) perturbations, whereas S\&P represents a non-gradient corruption mechanism; in addition, we define a mixed configuration that combines these perturbations within a single adversarial-augmentation scheme.

The experimental design follows a train and test matrix across perturbation types (see Figure \ref{fig:flow chart set up}). For each model family, we train on clean data and on adversarially augmented data produced by FGSM, PGD, S\&P, DeepFool and the mixed scheme, and then evaluate on both clean and perturbed test sets generated by each attack type. This setup supports evaluation of same-attack robustness and cross-attack generalisation. Performance is measured using ROC AUC, accuracy, precision, recall, and F1-score on clean data, alongside adversarial accuracy on perturbed inputs, with stratified 5-fold cross-validation to obtain reliable estimates.

\subsection{Dataset and Data Exploration}

\begin{figure}[thpb]
	\centering
	\begin{subfigure}[b]{0.47\textwidth}
		\centering
			\includegraphics[width=\textwidth]{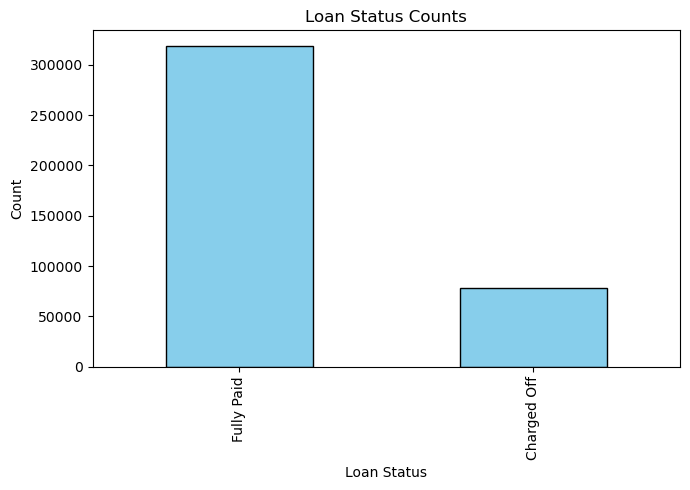}
		\caption
		{Distribution of the target variable (loan status).}    
		\label{fig:target-distribution}
	\end{subfigure}
	\hfill
	\begin{subfigure}[b]{0.47\textwidth}  
		\centering 
		\includegraphics[width=\textwidth]{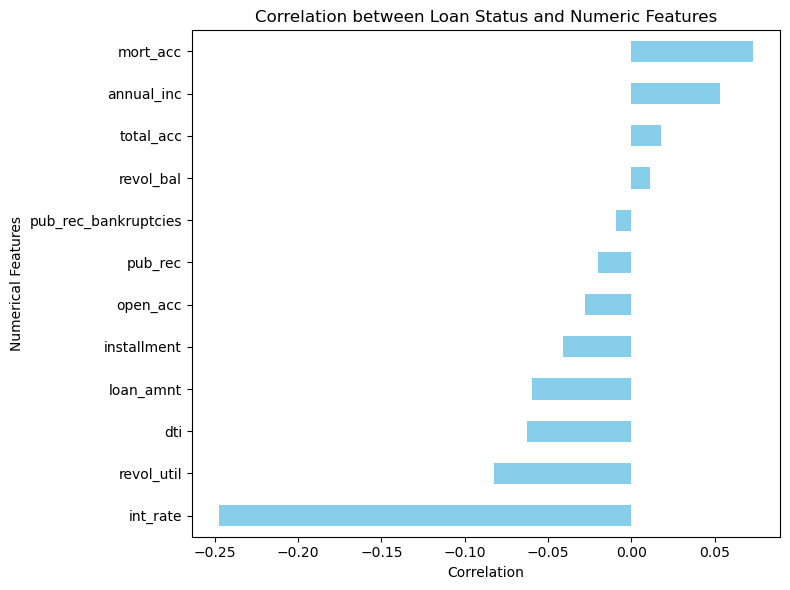}
		\caption{Correlation between relevant features and the target.}    
		\label{fig:corr-target}
	\end{subfigure}
	\caption{Target distribution and correction with relevant features in the dataset.} 
	\label{fig:target-info}
\end{figure}

We use a widely studied subset of the Lending Club loan dataset\footnote{https://www.kaggle.com/datasets/adarshsng/lending-club-loan-data-csv}, comprising nearly 400,000 observations and 28 features spanning borrower characteristics, loan terms, and credit history attributes (see Figures~\ref{fig:target-distribution}--\ref{fig:corr-target} and Table~\ref{tab:descriptive-stats})\footnote{The full dataset posed loading and preparation issues, given order of complexity of methods explored in the study; we therefore proceed with a curated subset that preserves the original schema and distributional properties.}. Consistent with prior work on consumer credit modelling \citep{van2025can, MachadoKarray2022_hybrid, zhang2020credit}, we include both numerical (e.g., loan\_amnt, int\_rate, dti) and categorical variables (e.g., term, grade, purpose). Exploratory analysis reveals a clear class imbalance: Fully Paid loans substantially outnumber Charged Off loans (Figure~\ref{fig:target-distribution}), which motivates the use of robust validation procedures and threshold-insensitive metrics (e.g., ROC AUC) in addition to accuracy. The descriptive statistics (Table~\ref{tab:descriptive-stats}) further highlight typical operational data issues, including heavy-tailed distributions and extreme values (e.g., dti and revol\_util), as well as limited missingness (notably in revol\_util).

These dataset characteristics are directly relevant for adversarial ML. Figure~\ref{fig:corr-target} indicates that several numerical features exhibit non-negligible associations with loan status, with int\_rate showing the strongest (negative) correlation and variables such as mort\_acc and annual\_inc showing positive correlations, while most other features are weakly correlated. From a robustness perspective, this suggests that adversarial perturbations targeted at a small subset of influential variables may disproportionately affect predictions, even if the overall linear correlation structure appears modest \citep{azzeh2018salt, naseem2024trans, kabir2025giids}. At the same time, the presence of outliers and implausible values underscores a practical challenge for AML on tabular credit data: unconstrained perturbations can easily generate unrealistic applicants, confounding robustness claims. This motivates the subsequent preprocessing and scaling choices, which aim to (i) stabilise model training and (ii) ensure that perturbed samples remain within a plausible numerical range when evaluating robustness under FGSM, PGD, S\&P, DeepFool and Mixed perturbations.

\begin{table}[ht]
  \centering
  \footnotesize
  \caption{Descriptive statistics of numerical features}
  \label{tab:descriptive-stats}
  \begin{tabularx}{\textwidth}{@{} l
    >{\raggedleft\arraybackslash}p{0.95cm}
    >{\raggedleft\arraybackslash}p{0.85cm}
    >{\raggedleft\arraybackslash}p{0.95cm}
    >{\raggedleft\arraybackslash}p{1.4cm}
    >{\raggedleft\arraybackslash}p{0.75cm}
    >{\raggedleft\arraybackslash}p{0.95cm}
    >{\raggedleft\arraybackslash}p{0.75cm}
    >{\raggedleft\arraybackslash}p{1.3cm}
    >{\raggedleft\arraybackslash}p{1cm}
    >{\raggedleft\arraybackslash}p{0.95cm}
    @{}}
    \toprule
              & Loan Amt & Int Rate & Install & Annual Inc & DTI & Open Acc & Pub Rec & Revol Bal & Revol Util & Total Acc \\
    \midrule
    Count     & 396\,030 & 396\,030 & 396\,030 & 396\,030 & 396\,030 & 396\,030 & 396\,030 & 396\,030 & 395\,754 & 396\,030 \\
    Mean      & 14\,113.89 & 13.63 & 431.84 & 74\,203.17 & 17.37 & 11.31 & 0.17 & 15\,844.53 & 53.79 & 25.41 \\
    STD       & 8\,357.44 & 4.47 & 250.72 & 61\,637.62 & 18.02 & 5.13 & 0.53 & 20\,591.83 & 24.45 & 11.89 \\
    Min       & 500 & 5.32 & 16.08 & 0 & 0 & 0 & 0 & 0 & 0 & 0 \\
    25\%      & 8\,000 & 10.49 & 250.33 & 45\,000 & 11.28 & 8 & 0 & 6\,025 & 35.80 & 17 \\
    Median    & 12\,000 & 13.33 & 375.43 & 64\,000 & 16.91 & 10 & 0 & 11\,181 & 54.80 & 24 \\
    75\%      & 20\,000 & 16.49 & 567.30 & 90\,000 & 22.98 & 14 & 0 & 19\,620 & 72.90 & 32 \\
    Max       & 40\,000 & 30.99 & 1\,533.81 & 8\,706\,582 & 9\,999 & 90 & 86 & 1\,743\,266 & 892.30 & 151 \\
    \bottomrule
  \end{tabularx}
\end{table}

\subsection{Preprocessing Pipeline}
\label{subsec:preproc}

We implement a reproducible preprocessing pipeline that converts the raw Lending Club extract into an analysis-ready table for model training and adversarial evaluation. The steps below consolidate the data cleaning and feature construction decisions used throughout the study \citep{van2025can, MachadoKarray2022_hybrid, zhang2020credit}:

\begin{itemize}
  \item Target definition and imbalance: The variable \texttt{loan\_status} is binarised into Fully Paid versus Charged Off. The resulting class imbalance (Figure \ref{fig:target-distribution}) motivates the use of class weighting during training (Section \ref{subsec:attacks}).
  \item Missing values: Six variables exhibit missingness: \texttt{emp\_title}, \texttt{emp\_length}, \texttt{title}, \texttt{revol\_util}, \texttt{mort\_acc}, and \texttt{pub\_rec\_bankruptcies}. We drop \texttt{emp\_title} due to high cardinality and more than 5\% missingness, and we drop \texttt{emp\_length} given its weak association with the target (Table \ref{tab:emp-length-charged}). We drop \texttt{title} as it largely overlaps with \texttt{purpose}. For \texttt{mort\_acc} (9.54\% missing), we impute using the conditional mean within \texttt{total\_acc} strata, leveraging their strong correlation (Figure \ref{fig:corr-target}) \citep{ZhangYu2024}. Rows with missing \texttt{revol\_util} (0.07\%) and \texttt{pub\_rec\_bankruptcies} (0.14\%) are removed.
  \item Categorical encoding: We apply binary encoding to \texttt{term} (36/60 months $\rightarrow$ \{0,1\}) and \texttt{initial\_list\_status} (w/f $\rightarrow$ \{0,1\}). We map \texttt{grade} and \texttt{sub\_grade} ordinally (A1--G5 $\rightarrow$ 1--35) and then drop \texttt{grade} and the raw \texttt{sub\_grade}. We apply one-hot encoding to \texttt{verification\_status}, \texttt{purpose}, \texttt{home\_ownership}, \texttt{application\_type}, and \texttt{zip\_code}.
  \item Leakage prevention: We drop \texttt{issue\_d}, which would not be available at decision time.
  \item Outlier handling: We remove implausible values using threshold filters informed by Table \ref{tab:descriptive-stats}, including anomalous values in \texttt{revol\_util}, to reduce leverage effects while preserving the bulk of the distribution.
  \item Feature scaling: We apply Min-Max scaling to $[0,1]$ for all continuous inputs to ensure comparability across features and to support consistent adversarial perturbation magnitudes.
\end{itemize}

\begin{table}[h]
  \centering
  \small
  \caption{Charged-off rates by employment length.}
  \label{tab:emp-length-charged}
  \begin{tabular}{@{}lcc@{}}
    \toprule
    \textbf{Employment length / loan status} & \textbf{Paid} & \textbf{Not paid} \\
    \midrule
    10+ years  & 0.8158 & 0.1842 \\
    9 years    & 0.7995 & 0.2005 \\
    8 years    & 0.8002 & 0.1998 \\
    7 years    & 0.8052 & 0.1948 \\
    6 years    & 0.8108 & 0.1891 \\
    5 years    & 0.8078 & 0.1922 \\
    4 years    & 0.8076 & 0.1923 \\
    3 years    & 0.8048 & 0.1952 \\
    2 years    & 0.8067 & 0.1932 \\
    1 year     & 0.8009 & 0.1991 \\
    $<1$ year   & 0.7931 & 0.2069 \\
    \bottomrule
  \end{tabular}
\end{table}

\begin{figure}[h]
  \centering
  \includegraphics[width=0.75\linewidth]{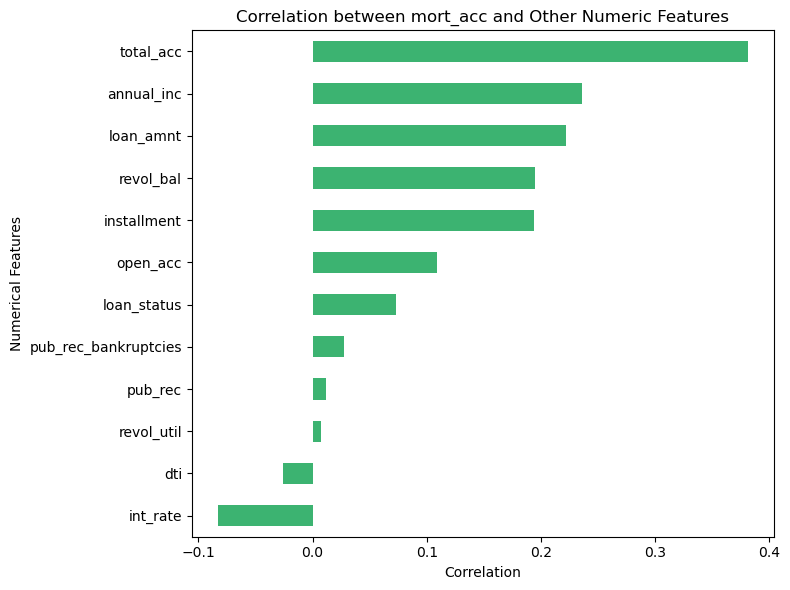}
  \caption{Correlation between \textit{mort\_acc} and other numeric features}
  \label{fig:mort numeric corr}
\end{figure}

\subsection{Model Specification, Training, Testing and Evaluating}
\label{subsec:attacks}

We split the dataset into 80\% training and 20\% test sets. To account for class imbalance, we use class weighting in the loss function to penalise minority-class errors more heavily, preserving the original class distribution and avoiding potential artefacts introduced by resampling \citep{van2025can, gouvea2007credit}. All reported results follow the same split logic and evaluation procedure to ensure comparability across models, training regimes, and attack conditions.

We evaluate three model families introduced in Section~\ref{sec:litreview}: logistic regression (linear baseline), a feed-forward neural network, and a transformer-based model for tabular data. Table ~\ref{tab:hyperparams_all_models} reports the hyperparameters for each architecture. The neural network uses batch normalisation and dropout in the input and hidden layers, a sigmoid output layer for binary classification, binary cross-entropy loss, the Adam optimiser, and early stopping to limit overfitting \citep{west2000neural}. Hyperparameters were selected through manual pilot tuning; the clean benchmark achieves a ROC AUC of 0.904, and we deliberately restrict exhaustive hyperparameter search to keep the focus on adversarial robustness rather than maximising clean-data performance.


\begin{table}[htbp]
  \centering
  \caption{Hyperparameters setting for the different models}
  \label{tab:hyperparams_all_models}
  \small
  \setlength{\tabcolsep}{6pt}
  \begin{tabular}{@{} l l l l @{}}
    \toprule
    \textbf{Hyperparameter} & \textbf{Neural network} & \textbf{FT-Transformer} & \textbf{Logistic regression} \\
    \midrule
    Hidden units            & \texttt{[150, 150, 150]} & --  & None \\
    Dropout rate(s)         & \texttt{[0.1, 0, 0.1, 0]} & 0.1 & None \\
    Activation function     & --                       & --  & Sigmoid \\
    Sequence length         & --                       & 4   & -- \\
    Embedding dimension     & --                       & 32  & -- \\
    Attention heads         & --                       & 2   & -- \\
    Key dimension           & --                       & 16  & -- \\
    Learning rate           & \texttt{1e-3}            & $10^{-3}$ & $10^{-3}$ \\
    Training epochs         & 20                       & 20  & 20 \\
    Batch size              & 32                       & 32  & 32 \\
    \bottomrule
  \end{tabular}
\end{table}

We focus on evasion attacks at inference time under two attacker knowledge regimes: white-box attacks with gradient access and black-box attacks without gradient access. The attack set follows Section~\ref{sec:litreview} and includes FGSM, PGD, DeepFool, and Salt-and-Pepper (S\&P) noise, with an additional mixed condition (part of the major contributions of this work) used for adversarial augmentation and testing:

\begin{itemize}
  \item FGSM and PGD: gradient-based white-box attacks that perturb inputs using the gradient sign (FGSM) or iterative projected updates (PGD) within an $\ell_\infty$ ball of radius $\varepsilon$ \citep{sarkar2018, hariprasad2025, HeOuyangWenLiuMoreira2025}. We also implement targeted variants that push charged-off cases toward the fully-paid class to reflect incentives in adversarial lending contexts \citep{sarkar2018, hariprasad2025, HeOuyangWenLiuMoreira2025}.
  \item Salt-and-Pepper (S\&P): a non-gradient perturbation that flips a proportion of feature values to extremes, approximating coarse manipulation or data quality shocks \citep{PelekisEtAl2025, azzeh2018salt}.
  \item DeepFool: a boundary-seeking adversarial attack that iteratively computes the minimal perturbation required to cross the model’s decision boundary under a local linear approximation \citep{moosavi2016deepfool}. Unlike FGSM and PGD, which operate within a predefined norm budget, DeepFool explicitly targets the closest decision boundary and therefore generates small, targeted perturbations that expose local geometric fragility in the classifier. Hence, not requiring attack strength parameter tuning.
\end{itemize}

To ensure that adversarial examples correspond to feasible applicant behaviour, we restrict all perturbations to features an applicant could plausibly manipulate: the continuous variables \texttt{annual\_inc}, \texttt{dti}, \texttt{revol\_bal}, and \texttt{revol\_util}, and the mutable categorical variables \texttt{purpose} and \texttt{zip\_code}. For the gradient-based attacks (FGSM, PGD, DeepFool), a binary mask zeroes the gradient of all immutable and remaining categorical features during the backward pass, so that only mutable features are updated and impossible feature states (e.g., fractional one-hot values) are never produced. For S\&P, min/max flips are applied only to the allowed continuous variables, and categorical corruption is implemented as a structurally valid swap that respects the mutual exclusivity of each one-hot block, deactivating the active category and activating a single alternative within the same block.

Attack strength is tuned to achieve meaningful but plausible degradation without inducing model collapse (Table~\ref{tab:attack-strength}). We select the mid-range setting $\varepsilon=0.1$ (FGSM/PGD) and \texttt{amount}=0.1 (S\&P). The sensitivity analysis in Table~\ref{tab:attack-strength} is computed on unconstrained perturbations to bracket the attack budget; once perturbations are restricted to mutable features as described above, the resulting degradation at $\varepsilon=0.1$ is milder, as reported in Section~\ref{sec:results}, while predictions remain stable.

\begin{table}[htbp]
  \centering
  \caption{Attack strength parameter tuning}
  \small
  \resizebox{\textwidth}{!}{%
    \begin{tabular}{c c c c c c c}
      \toprule
      $\varepsilon$/Amount 
        & Accuracy (FGSM) 
        & Accuracy (PGD) 
        & Accuracy (S\&P) 
        & ROC AUC (FGSM) 
        & ROC AUC (PGD) 
        & ROC AUC (S\&P) \\
      \midrule
      0.4   & 0.63 & 0.60 & 0.66               & 0.18 & 0.07 & 0.57 \\
      0.2   & 0.76 & 0.74 & 0.67               & 0.57 & 0.36 & 0.66 \\
      0.1   & 0.77 & 0.76 & 0.73\,(0.74)       & 0.78 & 0.74 & 0.74\,(0.75)\textsuperscript{w} \\
      0.05 & 0.78 & 0.78 & 0.77               & 0.84 & 0.83 & 0.81 \\
      0.02 & 0.81 & 0.81 & 0.82               & 0.86 & 0.86 & 0.89 \\
      \bottomrule
    \end{tabular}%
  }
  \label{tab:attack-strength}
\end{table}

Although the primary constraints on attack strength ($\varepsilon$ and amount of corruption) are selected based on the parameter tuning outlined in Table \ref{tab:attack-strength}, the internal parameters for iterative attacks are fixed to standard configurations established in the foundational literature. Table ~\ref{tab:attack_params_long} reports these internal parameters set for adversarial attacks.

For PGD, 40 iterations are employed with a step size of $\alpha=0.01$ \citep{MadryEtAl2017}. For DeepFool $\eta$ is set to 0.025 to guarantee decision boundary crossing, slightly increasing overshoot as proposed by \citep{moosavi2016deepfool}, and the number of max iterations is set to 100 \citep{Zhipengetal2025}.

\begin{table}[h!]
\centering
\caption{Hyperparameter settings for adversarial attacks}
\label{tab:attack_params_long}
\begin{tabular}{lll}
\hline
\textbf{Attack} & \textbf{Hyperparameter} & \textbf{Value} \\ \hline
\multirow{2}{*}{\textit{FGSM}} 
 & Attack budget ($\varepsilon$) & 0.1 \\
 & Target class & 1 (Fully Paid) \\ \hline
\multirow{4}{*}{\textit{PGD}} 
 & Attack budget ($\varepsilon$) & 0.1 \\
 & Step size ($\alpha$) & 0.01 \\
 & Iterations & 40 \\
 & Target class & 1 (Fully Paid) \\ \hline
\multirow{3}{*}{\textit{DeepFool}} 
 & Max iterations & 100 \\
 & Overshoot ($\eta$) & 0.025 \\
 & Target class & 1 (Fully Paid) \\ \hline
\multirow{2}{*}{\textit{Salt \& Pepper}} 
 & Corruption amount & 0.1 \\
 & Noise type & Min/Max flip \\ \hline
\end{tabular}
\end{table}

For each model family, we train one clean baseline and five adversarially augmented variants, using FGSM, PGD, S\&P, DeepFool and a mixed augmentation scheme. Adversarial augmentation is implemented by combining 80\% clean samples with 20\% adversarial samples generated from the corresponding attack method (Figure~\ref{fig:flow chart set up}). Each trained model is then evaluated on five test sets: clean, FGSM, PGD, DeepFool, and S\&P perturbations. This design enables within-regime comparisons (clean versus attacked test sets) and cross-regime comparisons (training on one perturbation type and testing on another), supporting a direct assessment of robustness transfer across attacks \citep{sarkar2018, hariprasad2025, HeOuyangWenLiuMoreira2025, PelekisEtAl2025}.

We report accuracy and ROC AUC, together with precision, recall, and F1-score. Given class imbalance, ROC AUC is used as the primary ranking metric \citep{HeOuyangWenLiuMoreira2025}. To quantify robustness, we additionally compute adversarial accuracy, defined as predictive performance under perturbed inputs \citep{PelekisEtAl2025}. Estimation stability is assessed via stratified 5-fold cross-validation, preserving class ratios within each fold \citep{van2025can}. Results are aggregated into a unified comparison table produced by a single evaluation script; confusion matrices and classification reports are generated for each model and test-set combination. The full experimental code and configuration files are available upon request and can be containerised for deterministic runs\footnote{Please see repository: https://github.com/Gijsn888/adversarialresearch}.

\section{Results and Discussions}
\label{sec:results}

This section compares model performance under clean and adversarial conditions, using the evaluation grid described in Section~\ref{subsec:attacks}. We report ROC AUC, accuracy, precision, recall, and F1 for both classes, together with the mean number of predicted non-defaults and defaults. Because misclassifying a default is typically more costly than misclassifying a non-default, we emphasise class-0 metrics (default detection), in particular Recall$_0$ and F1$_0$. Confusion matrices are generated for each configuration to support diagnostic interpretation beyond aggregate scores.

\subsection{``Clean" Baseline Performance}

Table~\ref{tab:model-comparison} reports clean-train/clean-test results for the three model families. All models achieve strong discrimination (ROC AUC 0.903--0.906). Logistic regression and the transformer achieve the highest accuracy (0.826), while all three models obtain similar ROC AUC (0.903--0.906). Across models, class-1 metrics are higher than class-0 metrics, indicating that non-defaults are easier to classify than defaults. The neural network exhibits the highest Recall$_0$ (0.765) and the lowest Precision$_0$ (0.534), reflecting a more conservative tendency to label cases as default, whereas the transformer attains the highest Precision$_0$ (0.542); logistic regression is comparable to the transformer (Recall$_0$ 0.756, Precision$_0$ 0.541).

\begin{table}[htbp]
  \centering
  \scriptsize
  \setlength{\tabcolsep}{3pt}
  \caption{Performance comparison of classification models evaluated on the same Clean Train--Clean Test split. Training and test identifiers are omitted as all metrics refer to an identical data partition.}
  \label{tab:model-comparison}
  \begin{tabular}{l *{10}{c}}
    \toprule
    Model 
      & ROC AUC & Accuracy 
      & Precision$_0$ & Recall$_0$ & F1$_0$ 
      & Precision$_1$ & Recall$_1$ & F1$_1$
      & Pred. Non-Defaults & Pred. Defaults \\
    \midrule

    Logistic Regression
      & 0.905 & 0.826
      & 0.541 & 0.756 & 0.631
      & 0.934 & 0.843 & 0.886
      & 45{,}367.2 & 17{,}232.2 \\
    Neural Network
      & 0.903 & 0.822
      & 0.534 & 0.765 & 0.628
      & 0.935 & 0.836 & 0.883
      & 44{,}911.4 & 17{,}688.0 \\
    Transformer
      & 0.906 & 0.826
      & 0.542 & 0.758 & 0.632
      & 0.934 & 0.843 & 0.886
      & 45{,}341.8 & 17{,}257.6 \\

\bottomrule
  \end{tabular}
\end{table}

\subsection{Impact of adversarial attacks on clean-trained models}

We next evaluate clean-trained models on perturbed test sets to measure vulnerability without any defensive training. Because perturbations are restricted to mutable features (Section~\ref{subsec:attacks}), the clean rows in Tables~\ref{tab:logit-adv-grid}, \ref{tab:nn-adv-grid}, and \ref{tab:transformer-adv-cv} show moderate rather than severe degradation. Logistic regression declines from ROC AUC 0.904 (clean) to 0.862 under FGSM and PGD and 0.870 under DeepFool. The neural network degrades comparably (ROC AUC 0.872--0.873 under FGSM/PGD and 0.878 under DeepFool), while for the transformer DeepFool is the most damaging attack (ROC AUC 0.856, versus 0.868 under FGSM/PGD). Across all three models, S\&P produces a similar decline (ROC AUC 0.855--0.864), indicating that the constrained attacks reduce discrimination consistently but without model collapse.

To assess the stability of these findings, we also ran an extensive cross-validation study using 1-, 2-, 3-, 4-, and 5-fold stratified cross-validation across the same clean-train/attacked-test settings. \ref{appendixA} reports the fold-averaged metrics for all cross-validation experiments, summarised separately for A1 (logistic regression), A2 (neural network), and A3 (transformer). These additional results confirm that the observed vulnerability patterns are not driven by a single split and remain consistent across different validation granularities.

The direction of the effect is clearer when examining predicted class counts. Table~\ref{tab:clean-train-predictions} shows that, for all three models, FGSM and PGD reduce the number of predicted defaults relative to the clean test set (e.g., logistic regression from 17{,}232.2 to 15{,}946.8; the neural network from 17{,}688.0 to 16{,}617.2 under FGSM). This is consistent with targeted attacks that aim to move default cases toward the non-default class. In contrast, S\&P increases the number of predicted defaults for all models and reduces both class-0 and class-1 performance, which is consistent with non-targeted corruption that broadly disrupts feature values rather than systematically inducing class flips.

\begin{table}[htbp]
\centering
\scriptsize
\setlength{\tabcolsep}{2.5pt}
\caption{Five fold evaluation of the Logistic Regression model trained and tested under different adversarial attack configurations.}
\label{tab:logit-adv-grid}
\resizebox{\textwidth}{!}{%
\begin{tabular}{l l c c c c c c c c c c}
\toprule
Training Set & Test Set &
ROC AUC & Accuracy &
Prec$_0$ & Prec$_1$ &
Rec$_0$ & Rec$_1$ &
F1$_0$ & F1$_1$ &
Pred ND & Pred D \\
\midrule

Clean & Clean
& 0.904 & 0.823
& 0.535 & 0.934
& 0.758 & 0.839
& 0.628 & 0.884
& 45146 & 17453 \\
Clean & FGSM
& 0.862 & 0.801
& 0.497 & 0.907
& 0.650 & 0.839
& 0.563 & 0.872
& 46481 & 16118 \\
Clean & PGD
& 0.862 & 0.801
& 0.497 & 0.907
& 0.650 & 0.839
& 0.563 & 0.872
& 46481 & 16118 \\
Clean & S\&P
& 0.861 & 0.793
& 0.483 & 0.926
& 0.737 & 0.806
& 0.583 & 0.862
& 43782 & 18817 \\
Clean & DeepFool
& 0.870 & 0.784
& 0.461 & 0.887
& 0.562 & 0.839
& 0.507 & 0.862
& 47557 & 15042 \\

\midrule

FGSM & Clean
& 0.898 & 0.831
& 0.554 & 0.924
& 0.710 & 0.860
& 0.623 & 0.891
& 46820 & 15779 \\
FGSM & FGSM
& 0.947 & 0.862
& 0.603 & 0.964
& 0.868 & 0.860
& 0.712 & 0.909
& 44874 & 17725 \\
FGSM & PGD
& 0.947 & 0.862
& 0.603 & 0.964
& 0.868 & 0.860
& 0.712 & 0.909
& 44874 & 17725 \\
FGSM & S\&P
& 0.855 & 0.802
& 0.497 & 0.915
& 0.687 & 0.830
& 0.577 & 0.870
& 45555 & 17044 \\
FGSM & DeepFool
& 0.909 & 0.837
& 0.566 & 0.932
& 0.745 & 0.860
& 0.643 & 0.895
& 46388 & 16211 \\

\midrule

PGD & Clean
& 0.898 & 0.817
& 0.526 & 0.930
& 0.745 & 0.835
& 0.616 & 0.880
& 45129 & 17470 \\
PGD & FGSM
& 0.948 & 0.848
& 0.572 & 0.971
& 0.899 & 0.835
& 0.699 & 0.898
& 43234 & 19365 \\
PGD & PGD
& 0.948 & 0.848
& 0.572 & 0.971
& 0.899 & 0.835
& 0.699 & 0.898
& 43234 & 19365 \\
PGD & S\&P
& 0.855 & 0.789
& 0.476 & 0.922
& 0.722 & 0.805
& 0.573 & 0.859
& 43902 & 18697 \\
PGD & DeepFool
& 0.910 & 0.822
& 0.533 & 0.936
& 0.768 & 0.835
& 0.629 & 0.883
& 44848 & 17751 \\

\midrule

S\&P & Clean
& 0.903 & 0.825
& 0.539 & 0.933
& 0.753 & 0.842
& 0.628 & 0.885
& 45380 & 17219 \\
S\&P & FGSM
& 0.864 & 0.804
& 0.502 & 0.908
& 0.650 & 0.842
& 0.567 & 0.874
& 46649 & 15950 \\
S\&P & PGD
& 0.864 & 0.804
& 0.502 & 0.908
& 0.650 & 0.842
& 0.567 & 0.874
& 46649 & 15950 \\
S\&P & S\&P
& 0.863 & 0.795
& 0.486 & 0.924
& 0.729 & 0.811
& 0.584 & 0.864
& 44114 & 18485 \\
S\&P & DeepFool
& 0.861 & 0.794
& 0.482 & 0.895
& 0.599 & 0.842
& 0.534 & 0.868
& 47280 & 15319 \\

\midrule

DeepFool & Clean
& 0.901 & 0.814
& 0.519 & 0.935
& 0.765 & 0.826
& 0.619 & 0.877
& 44440 & 18159 \\
DeepFool & FGSM
& 0.918 & 0.824
& 0.534 & 0.947
& 0.813 & 0.826
& 0.645 & 0.883
& 43855 & 18744 \\
DeepFool & PGD
& 0.918 & 0.824
& 0.534 & 0.947
& 0.813 & 0.826
& 0.645 & 0.883
& 43855 & 18744 \\
DeepFool & S\&P
& 0.858 & 0.785
& 0.470 & 0.926
& 0.741 & 0.795
& 0.576 & 0.856
& 43168 & 19431 \\
DeepFool & DeepFool
& 0.904 & 0.815
& 0.520 & 0.936
& 0.768 & 0.826
& 0.621 & 0.878
& 44400 & 18199 \\

\midrule

Mixed & Clean
& 0.900 & 0.820
& 0.531 & 0.931
& 0.746 & 0.838
& 0.620 & 0.882
& 45262 & 17337 \\
Mixed & FGSM
& 0.929 & 0.837
& 0.558 & 0.954
& 0.834 & 0.838
& 0.669 & 0.892
& 44178 & 18421 \\
Mixed & PGD
& 0.929 & 0.837
& 0.558 & 0.954
& 0.834 & 0.838
& 0.669 & 0.892
& 44178 & 18421 \\
Mixed & S\&P
& 0.858 & 0.791
& 0.480 & 0.922
& 0.722 & 0.808
& 0.577 & 0.861
& 44060 & 18539 \\
Mixed & DeepFool
& 0.906 & 0.823
& 0.535 & 0.934
& 0.759 & 0.838
& 0.628 & 0.884
& 45102 & 17497 \\

\bottomrule
\end{tabular}}
\end{table}

\begin{table}[htbp]
\centering
\scriptsize
\setlength{\tabcolsep}{2.5pt}
\caption{Five fold evaluation of the Neural Network model trained and tested under different adversarial attack configurations.}
\label{tab:nn-adv-grid}
\resizebox{\textwidth}{!}{%
\begin{tabular}{l l c c c c c c c c c c}
\toprule
Training Set & Test Set &
ROC AUC & Accuracy &
Prec$_0$ & Prec$_1$ &
Rec$_0$ & Rec$_1$ &
F1$_0$ & F1$_1$ &
Pred ND & Pred D \\
\midrule

Clean & Clean
& 0.901 & 0.815
& 0.520 & 0.937
& 0.773 & 0.825
& 0.622 & 0.877
& 44274 & 18325 \\
Clean & FGSM
& 0.873 & 0.800
& 0.494 & 0.918
& 0.698 & 0.825
& 0.579 & 0.869
& 45205 & 17394 \\
Clean & PGD
& 0.872 & 0.800
& 0.494 & 0.917
& 0.697 & 0.825
& 0.578 & 0.869
& 45212 & 17387 \\
Clean & S\&P
& 0.855 & 0.786
& 0.472 & 0.928
& 0.748 & 0.795
& 0.579 & 0.856
& 43073 & 19526 \\
Clean & DeepFool
& 0.878 & 0.810
& 0.513 & 0.931
& 0.751 & 0.825
& 0.610 & 0.875
& 44545 & 18054 \\

\midrule

FGSM & Clean
& 0.897 & 0.848
& 0.601 & 0.920
& 0.684 & 0.888
& 0.640 & 0.904
& 48555 & 14044 \\
FGSM & FGSM
& 0.952 & 0.880
& 0.651 & 0.959
& 0.847 & 0.888
& 0.736 & 0.923
& 46549 & 16050 \\
FGSM & PGD
& 0.954 & 0.882
& 0.653 & 0.961
& 0.854 & 0.888
& 0.740 & 0.923
& 46456 & 16143 \\
FGSM & S\&P
& 0.852 & 0.818
& 0.531 & 0.912
& 0.665 & 0.856
& 0.591 & 0.883
& 47154 & 15445 \\
FGSM & DeepFool
& 0.891 & 0.837
& 0.580 & 0.907
& 0.629 & 0.888
& 0.604 & 0.898
& 49238 & 13361 \\

\midrule

PGD & Clean
& 0.895 & 0.835
& 0.565 & 0.925
& 0.713 & 0.865
& 0.631 & 0.894
& 47033 & 15566 \\
PGD & FGSM
& 0.954 & 0.867
& 0.614 & 0.965
& 0.873 & 0.865
& 0.721 & 0.913
& 45064 & 17535 \\
PGD & PGD
& 0.959 & 0.870
& 0.618 & 0.969
& 0.888 & 0.865
& 0.729 & 0.914
& 44885 & 17714 \\
PGD & S\&P
& 0.854 & 0.806
& 0.506 & 0.917
& 0.692 & 0.835
& 0.585 & 0.874
& 45753 & 16846 \\
PGD & DeepFool
& 0.895 & 0.830
& 0.556 & 0.919
& 0.688 & 0.865
& 0.615 & 0.891
& 47341 & 15258 \\

\midrule

S\&P & Clean
& 0.898 & 0.819
& 0.529 & 0.934
& 0.758 & 0.834
& 0.623 & 0.881
& 44915 & 17684 \\
S\&P & FGSM
& 0.863 & 0.801
& 0.497 & 0.911
& 0.667 & 0.834
& 0.569 & 0.871
& 46040 & 16559 \\
S\&P & PGD
& 0.862 & 0.801
& 0.496 & 0.910
& 0.665 & 0.834
& 0.568 & 0.871
& 46064 & 16535 \\
S\&P & S\&P
& 0.873 & 0.798
& 0.492 & 0.925
& 0.732 & 0.814
& 0.588 & 0.866
& 44244 & 18355 \\
S\&P & DeepFool
& 0.872 & 0.810
& 0.512 & 0.921
& 0.709 & 0.834
& 0.595 & 0.876
& 45520 & 17079 \\

\midrule

DeepFool & Clean
& 0.900 & 0.806
& 0.504 & 0.939
& 0.787 & 0.810
& 0.615 & 0.870
& 43363 & 19236 \\
DeepFool & FGSM
& 0.888 & 0.801
& 0.496 & 0.933
& 0.761 & 0.810
& 0.601 & 0.867
& 43676 & 18923 \\
DeepFool & PGD
& 0.889 & 0.801
& 0.497 & 0.933
& 0.764 & 0.810
& 0.602 & 0.867
& 43649 & 18950 \\
DeepFool & S\&P
& 0.853 & 0.778
& 0.462 & 0.930
& 0.760 & 0.783
& 0.574 & 0.850
& 42304 & 20295 \\
DeepFool & DeepFool
& 0.901 & 0.805
& 0.503 & 0.939
& 0.784 & 0.810
& 0.613 & 0.870
& 43396 & 19203 \\

\midrule

Mixed & Clean
& 0.899 & 0.834
& 0.562 & 0.926
& 0.719 & 0.862
& 0.631 & 0.893
& 46815 & 15784 \\
Mixed & FGSM
& 0.912 & 0.842
& 0.574 & 0.935
& 0.757 & 0.862
& 0.653 & 0.897
& 46355 & 16244 \\
Mixed & PGD
& 0.915 & 0.843
& 0.577 & 0.937
& 0.765 & 0.862
& 0.658 & 0.898
& 46247 & 16352 \\
Mixed & S\&P
& 0.867 & 0.807
& 0.508 & 0.918
& 0.696 & 0.835
& 0.588 & 0.874
& 45700 & 16899 \\
Mixed & DeepFool
& 0.898 & 0.831
& 0.557 & 0.923
& 0.705 & 0.862
& 0.622 & 0.892
& 46991 & 15608 \\

\bottomrule
\end{tabular}}
\end{table}

\begin{table}[htbp]
\centering
\scriptsize
\setlength{\tabcolsep}{2.5pt}
\caption{Five fold evaluation of the Transformer model trained and tested under different adversarial attack combinations. For each training configuration.}
\label{tab:transformer-adv-cv}
\resizebox{\textwidth}{!}{%
\begin{tabular}{l l c c c c c c c c c c}
\toprule
Training Set & Test Set &
ROC AUC & Accuracy &
Prec$_0$ & Prec$_1$ &
Rec$_0$ & Rec$_1$ &
F1$_0$ & F1$_1$ &
Pred ND & Pred D \\
\midrule

Clean & Clean
& 0.905 & 0.824
& 0.538 & 0.934
& 0.759 & 0.840
& 0.630 & 0.885
& 45216 & 17383 \\
Clean & FGSM
& 0.868 & 0.804
& 0.502 & 0.909
& 0.656 & 0.840
& 0.569 & 0.873
& 46486 & 16113 \\
Clean & PGD
& 0.868 & 0.804
& 0.502 & 0.909
& 0.656 & 0.840
& 0.569 & 0.873
& 46486 & 16113 \\
Clean & S\&P
& 0.864 & 0.795
& 0.486 & 0.927
& 0.739 & 0.808
& 0.586 & 0.863
& 43862 & 18737 \\
Clean & DeepFool
& 0.856 & 0.774
& 0.437 & 0.874
& 0.505 & 0.840
& 0.469 & 0.857
& 48342 & 14257 \\

\midrule

FGSM & Clean
& 0.901 & 0.846
& 0.596 & 0.920
& 0.684 & 0.886
& 0.637 & 0.903
& 48443 & 14156 \\
FGSM & FGSM
& 0.980 & 0.898
& 0.670 & 0.985
& 0.944 & 0.886
& 0.784 & 0.933
& 45233 & 17366 \\
FGSM & PGD
& 0.980 & 0.898
& 0.670 & 0.985
& 0.945 & 0.886
& 0.784 & 0.933
& 45232 & 17367 \\
FGSM & S\&P
& 0.858 & 0.816
& 0.526 & 0.912
& 0.664 & 0.853
& 0.587 & 0.882
& 47038 & 15561 \\
FGSM & DeepFool
& 0.894 & 0.839
& 0.582 & 0.911
& 0.646 & 0.886
& 0.612 & 0.898
& 48911 & 13688 \\

\midrule

PGD & Clean
& 0.902 & 0.828
& 0.546 & 0.930
& 0.739 & 0.849
& 0.628 & 0.888
& 45920 & 16679 \\
PGD & FGSM
& 0.976 & 0.868
& 0.606 & 0.984
& 0.943 & 0.849
& 0.738 & 0.912
& 43399 & 19200 \\
PGD & PGD
& 0.976 & 0.868
& 0.606 & 0.984
& 0.943 & 0.849
& 0.738 & 0.912
& 43399 & 19200 \\
PGD & S\&P
& 0.861 & 0.798
& 0.491 & 0.922
& 0.716 & 0.818
& 0.582 & 0.867
& 44598 & 18001 \\
PGD & DeepFool
& 0.911 & 0.826
& 0.543 & 0.927
& 0.729 & 0.849
& 0.622 & 0.887
& 46046 & 16553 \\

\midrule

S\&P & Clean
& 0.904 & 0.834
& 0.561 & 0.927
& 0.725 & 0.861
& 0.633 & 0.893
& 46662 & 15937 \\
S\&P & FGSM
& 0.864 & 0.814
& 0.524 & 0.903
& 0.625 & 0.861
& 0.570 & 0.882
& 47901 & 14698 \\
S\&P & PGD
& 0.864 & 0.814
& 0.524 & 0.903
& 0.625 & 0.861
& 0.570 & 0.882
& 47901 & 14698 \\
S\&P & S\&P
& 0.879 & 0.813
& 0.518 & 0.920
& 0.703 & 0.840
& 0.597 & 0.878
& 45876 & 16723 \\
S\&P & DeepFool
& 0.853 & 0.783
& 0.451 & 0.868
& 0.467 & 0.861
& 0.459 & 0.864
& 49847 & 12752 \\

\midrule

DeepFool & Clean
& 0.903 & 0.835
& 0.564 & 0.927
& 0.722 & 0.863
& 0.633 & 0.894
& 46818 & 15781 \\
DeepFool & FGSM
& 0.894 & 0.833
& 0.560 & 0.924
& 0.711 & 0.863
& 0.627 & 0.893
& 46951 & 15648 \\
DeepFool & PGD
& 0.894 & 0.833
& 0.560 & 0.924
& 0.711 & 0.863
& 0.627 & 0.893
& 46951 & 15648 \\
DeepFool & S\&P
& 0.864 & 0.805
& 0.504 & 0.920
& 0.706 & 0.830
& 0.588 & 0.872
& 45331 & 17268 \\
DeepFool & DeepFool
& 0.907 & 0.832
& 0.559 & 0.923
& 0.707 & 0.863
& 0.624 & 0.892
& 46999 & 15600 \\

\midrule

Mixed & Clean
& 0.903 & 0.825
& 0.539 & 0.932
& 0.750 & 0.843
& 0.627 & 0.885
& 45467 & 17132 \\
Mixed & FGSM
& 0.979 & 0.865
& 0.598 & 0.987
& 0.954 & 0.843
& 0.736 & 0.909
& 42945 & 19654 \\
Mixed & PGD
& 0.979 & 0.865
& 0.598 & 0.987
& 0.954 & 0.843
& 0.736 & 0.909
& 42945 & 19654 \\
Mixed & S\&P
& 0.875 & 0.799
& 0.493 & 0.924
& 0.728 & 0.816
& 0.588 & 0.867
& 44379 & 18220 \\
Mixed & DeepFool
& 0.917 & 0.826
& 0.541 & 0.934
& 0.756 & 0.843
& 0.631 & 0.886
& 45390 & 17209 \\

\bottomrule
\end{tabular}}
\end{table}

\begin{table}[htbp]
\centering
\small
\setlength{\tabcolsep}{6pt}
\caption{Mean predicted non-defaults and defaults for models trained on clean data and evaluated under different test-time perturbations. Values represent fold-averaged predictions.}
\label{tab:clean-train-predictions}
\begin{tabular}{l l l c c}
\toprule
Model & Training Set & Test Set &
Predicted Non-Defaults & Predicted Defaults \\
\midrule

Logistic Regression & Clean Train & Clean Test     & 45{,}367.2 & 17{,}232.2 \\
Logistic Regression & Clean Train & DeepFool Test     & 47{,}770.4 & 14{,}829.0 \\
Logistic Regression & Clean Train & FGSM Test     & 46{,}652.6 & 15{,}946.8 \\
Logistic Regression & Clean Train & PGD Test     & 46{,}652.6 & 15{,}946.8 \\
Logistic Regression & Clean Train & S\&P Test     & 43{,}903.8 & 18{,}695.6 \\

\midrule

Neural Network & Clean Train & Clean Test     & 44{,}911.4 & 17{,}688.0 \\
Neural Network & Clean Train & DeepFool Test     & 46{,}397.4 & 16{,}202.0 \\
Neural Network & Clean Train & FGSM Test     & 45{,}982.2 & 16{,}617.2 \\
Neural Network & Clean Train & PGD Test     & 45{,}983.8 & 16{,}615.6 \\
Neural Network & Clean Train & S\&P Test     & 43{,}479.6 & 19{,}119.8 \\

\midrule

Transformer & Clean Train & Clean Test     & 45{,}341.8 & 17{,}257.6 \\
Transformer & Clean Train & DeepFool Test     & 47{,}260.8 & 15{,}338.6 \\
Transformer & Clean Train & FGSM Test     & 46{,}553.2 & 16{,}046.2 \\
Transformer & Clean Train & PGD Test     & 46{,}552.2 & 16{,}047.2 \\
Transformer & Clean Train & S\&P Test     & 43{,}923.4 & 18{,}676.0 \\

\bottomrule
\end{tabular}
\end{table}

\subsection{Adversarial training and cross-attack generalisation}

Adversarial training substantially improves robustness when the training and test perturbations match, especially for gradient-based attacks. For all model families, training on FGSM or PGD raises ROC AUC on FGSM and PGD test sets well above the clean-trained baseline. For example, the neural network reaches ROC AUC 0.952--0.959 on FGSM/PGD tests, with accuracy 0.867--0.882 (Table~\ref{tab:nn-adv-grid}), and the transformer reaches up to ROC AUC 0.980 on FGSM/PGD tests, with accuracy 0.868--0.898 (Table~\ref{tab:transformer-adv-cv}). Logistic regression shows the same pattern, reaching ROC AUC 0.947--0.948 on matched tests (Table~\ref{tab:logit-adv-grid}). Robustness also transfers well between FGSM and PGD: models trained on one perform almost identically when tested on the other, reflecting their shared gradient-based structure.

The improvement of ROC AUC for models trained and tested on matching FGSM/PGD requires specific interpretation. Since these attacks apply a fixed-magnitude transformation ($\varepsilon = 0.1$) to the mutable features, they introduce a partial structural shift in feature space that the adversarially trained model learns to recognise, making perturbed defaults easier to separate. Because the perturbations are confined to a small set of mutable features rather than the full input, this shift is limited, which is why matched training improves discrimination substantially but does not converge to a perfect score.

By contrast, DeepFool training improves robustness but by a smaller margin than FGSM/PGD. Since DeepFool computes the minimal perturbations required to cross the decision boundary, the adversarial examples remain close to the true class boundary. The resulting improvement therefore reflects genuine boundary hardening rather than recognition of a fixed perturbation pattern. 

Cross-attack generalisation to S\&P is weaker than within the gradient family, but it is stable rather than catastrophic. Gradient-based adversarial training does not markedly improve S\&P robustness: for all three models, FGSM/PGD-trained models attain ROC AUC around 0.852--0.861 and accuracy around 0.79--0.82 on S\&P tests, close to the clean-trained models. S\&P-based training gives the strongest S\&P performance (ROC AUC 0.863 for logistic regression, 0.873 for the neural network, and 0.879 for the transformer) but does not strengthen resistance to FGSM/PGD, consistent with S\&P corruption having a different structure from the gradient-based attacks.

The mixed training regime provides the most balanced results for the neural network and transformer. It maintains clean performance while achieving strong robustness on FGSM/PGD and improved resilience under DeepFool. For the transformer, mixed training achieves accuracy 0.825 on the clean test set and 0.826 on the DeepFool test set, while reaching accuracy 0.865 and ROC AUC 0.979 on FGSM/PGD tests, essentially matching the best single-attack model (Table~\ref{tab:transformer-adv-cv}). For the neural network, mixed training improves performance under FGSM/PGD relative to clean training while maintaining stable behaviour under S\&P (Table~\ref{tab:nn-adv-grid}). These results support the use of multi-attack augmentation when the threat model is uncertain or heterogeneous. As an external robustness check, we reproduce this train--test grid on an independent P2P dataset (Prosper) in \ref{appendixB}; the same ordering, matched gradient-based training yields the largest gains, FGSM and PGD remain mutually interchangeable, cross-family transfer to S\&P is weak, and mixed training is the most balanced, holds despite that dataset's lower overall discrimination.

A common concern is that adversarial training may reduce clean-data performance. In our results, this trade-off is limited for the neural network and transformer, and sometimes favourable: adversarially trained variants remain close to the clean benchmarks on the clean test set (e.g., the transformer under FGSM training reaches accuracy 0.846 on the clean test set, slightly above the 0.824 of the clean-trained transformer, while PGD and mixed training reach 0.828 and 0.825, Table~\ref{tab:transformer-adv-cv}). Logistic regression shows smaller changes on clean tests and benefits less from adversarial training overall. Overall, the results suggest that, under the selected augmentation ratio and perturbation strengths, improved robustness does not require sacrificing clean performance for the higher-capacity models.

\subsection{Comparison with the literature}

The findings align with the broader AML literature in which adversarial training is most effective when the training threat model matches the evaluation threat model, particularly for gradient-based attacks \citep{MadryEtAl2017,PelekisEtAl2025}. They also complement work examining adversarial vulnerability and defence mechanisms for credit risk and P2P settings \citep{sarkar2018,schwab2025mitigating,straus2025explaining}. The main contribution of this study is the systematic train--test grid across multiple perturbation families, which makes the limits of transfer explicit: robustness under FGSM/PGD does not automatically imply robustness to non-gradient corruption, while mixed training provides a more reliable default when the attacker strategy is unknown.

\begin{table}[ht]
\centering
\caption{Results study \citep{ZhangEtAl2025}}
\begin{tabular}{lcccc}
\toprule
\textbf{Model} & \textbf{Accuracy} & \textbf{Recall} & \textbf{Precision} & \textbf{F1 Score} \\
\midrule
SVM       & 0.83 & 0.88 & 0.84 & 0.89 \\
BPnetwork & 0.88 & 0.93 & 0.88 & 0.94 \\
RNN       & 0.90 & 0.96 & 0.90 & 0.95 \\
LSTM      & 0.92 & 0.97 & 0.93 & 0.96 \\
GANs      & 0.96 & 1.00 & 0.97 & 0.97 \\
\bottomrule
\end{tabular}
\label{tab:zhang-results}
\end{table}

The study in Table~\ref{tab:zhang-results} illustrates that augmenting training data with synthetic scenarios can improve predictive performance in credit risk settings \citep{ZhangEtAl2025}. While that work does not evaluate targeted evasion attacks, the broader message is consistent with our results: model improvements from augmented data depend on whether the augmented samples reflect the stress conditions a deployed system will face. Our findings extend this perspective by showing that adversarial augmentation produces large gains under the corresponding attack family, and that mixed augmentation is a practical route to stronger resilience under heterogeneous perturbations.

\subsection{Managerial implications}
\label{subsec:managerial}

First, the results indicate that clean benchmark performance is not a sufficient proxy for operational reliability. Models with strong clean ROC AUC and accuracy can exhibit sharp degradation under targeted perturbations, including a systematic reduction in predicted defaults under FGSM/PGD. For P2P platforms and lenders, this supports incorporating adversarial stress testing into model validation and ongoing monitoring, using both gradient-based attacks (to represent strategic manipulation) and non-gradient perturbations (to represent data quality shocks and coarse manipulation). Monitoring should prioritise default-sensitive metrics (Recall$_0$, F1$_0$) under stress, since these align more directly with credit losses than overall accuracy.

Second, the train--test grid shows that single-attack defences can create a false sense of security. FGSM- and PGD-trained models are highly robust to FGSM/PGD tests, yet this robustness does not fully transfer to non-gradient corruption such as S\&P. A practical management response is to adopt mixed or multi-attack training as a default when the attacker strategy is uncertain, especially for higher-capacity models that maintain strong clean performance while gaining robustness across multiple threat conditions. This supports a risk-based hardening strategy: deploy mixed training for high-impact decision segments (large exposures, thin-file borrowers, automated approvals) and maintain periodic re-training as manipulation patterns evolve.

Third, robustness should be paired with operational controls. Because non-gradient perturbations can represent noisy or implausible inputs, lenders can reduce exposure by implementing plausibility checks, consistency rules, and verification against external sources where available, especially for features that are influential in the model. Combined with adversarial training, these controls can reduce both economic loss and model-governance risk by improving stability, auditability, and compliance alignment for high-stakes credit decision systems.

\section{Conclusion}
\label{sec:conclusion}

This paper examined whether adversarial machine learning can strengthen the robustness of credit risk models in peer-to-peer lending. Motivated by the risk that applicants may manipulate input information to obtain favourable lending decisions, we compared standard training with adversarial training on real-world Lending Club data. The study evaluated three model families commonly used for tabular credit scoring (logistic regression, a feed-forward neural network, and a transformer-based model) under clean conditions and under adversarially perturbed test sets. The central objective was to assess how adversarially trained models compare with traditionally trained models when predictions are made in environments that include targeted and non-targeted input manipulation.

Across models, the results show that adversarial training substantially improves robustness when the attack mechanism used at test time matches the perturbation family used during training. In particular, models trained with gradient-based perturbations show strong resilience to gradient-based attacks, with large improvements in discrimination and default-detection metrics relative to clean-trained baselines evaluated under attack. These robustness gains are achieved without a material deterioration in clean-test performance for the higher-capacity models, suggesting that robustness can be strengthened while maintaining predictive usefulness in normal operating conditions. In contrast, Salt-and-Pepper perturbations behave as coarse, non-gradient corruption and yield smaller robustness gains than the gradient-based attacks when used for training, and this robustness transfers less to the other attack families.

The main contributions of this work are threefold. First, it provides an empirical robustness benchmark for credit scoring on a large P2P dataset using a systematic train--test evaluation grid across multiple attack families. Second, it compares robustness behaviour across three distinct model classes, showing that robustness profiles differ by architecture and by perturbation mechanism, and that strong clean accuracy does not guarantee resilience under manipulation. Third, it demonstrates the value of hybrid, mixed training strategies for improving robustness under heterogeneous conditions, which is closer to realistic deployment where attacker behaviour may vary and evolve. Together, these contributions help bridge the gap between AML methods developed in other domains and the practical requirements of robust financial decision systems.

\subsection{Limitations and future research}
\label{subsec:limitations_future}

This study has several limitations. First, the experiments rely on a curated subset of the Lending Club dataset and a fixed preprocessing pipeline; while this supports reproducibility and controlled comparison, the results may not fully generalise to other lenders, time periods, feature sets, or operational constraints. To probe external validity, \ref{appendixB} replicates the entire pipeline, the four attacks, the mutable-feature constraints, and the mixed regime, on an independent P2P dataset (Prosper). The qualitative robustness and cross-attack transfer patterns reported above are preserved on this structurally different and smaller subset, even though its absolute discrimination is markedly lower, which reinforces rather than removes the need for lender-specific validation. Second, the adversarial attacks applied here are stylised approximations of manipulation in tabular data. Although FGSM/PGD provide controlled gradient-based perturbations and S\&P represents coarse corruption, these methods do not fully encode feasibility constraints, behavioural incentives, or institutional verification processes that shape real borrower fraud.

Future research should therefore move toward threat models that reflect plausible manipulation strategies in credit applications. This includes constraint-aware attacks that respect feature semantics (e.g., monotonicity, discreteness, and budgeted changes), stronger black-box attacks tailored to tabular finance settings, and scenario-based evaluation that couples AML with domain rules and verification signals. It would also be valuable to study robustness under distribution shift, temporal drift, and policy changes, and to integrate robustness assessment into model governance frameworks with clear decision thresholds and cost-sensitive objectives. Finally, evaluating robustness alongside explainability and compliance requirements, including post-deployment monitoring and incident response, would provide a more complete blueprint for deploying resilient credit risk models in practice.

\bibliography{cas-refs}

\clearpage

\appendix
\section{Cross-validation results}
\label{appendixA}

\begin{table}[htbp]
\centering
\tiny
\setlength{\tabcolsep}{2.5pt}
\caption{Mean performance of the Logistic Regression model across cross-validation folds under different adversarial training and testing configurations. All values represent fold-averaged metrics and are rounded to three decimal digits.}
\label{tab:logit-mean-adv}
\resizebox{0.85\textwidth}{!}{%
\begin{tabular}{l l c c c c c c c c c c}
\toprule
Training Set & Test Set &
ROC AUC & Accuracy &
Prec$_0$ & Prec$_1$ &
Rec$_0$ & Rec$_1$ &
F1$_0$ & F1$_1$ &
Pred ND & Pred D \\
\midrule

Clean & Clean
& 0.905 & 0.826
& 0.541 & 0.934
& 0.756 & 0.843
& 0.631 & 0.886
& 45367 & 17232 \\
Clean & DeepFool
& 0.872 & 0.787
& 0.465 & 0.887
& 0.561 & 0.843
& 0.508 & 0.864
& 47770 & 14829 \\
Clean & FGSM
& 0.864 & 0.805
& 0.504 & 0.908
& 0.652 & 0.843
& 0.569 & 0.874
& 46653 & 15947 \\
Clean & PGD
& 0.864 & 0.805
& 0.504 & 0.908
& 0.652 & 0.843
& 0.569 & 0.874
& 46653 & 15947 \\
Clean & S\&P
& 0.862 & 0.795
& 0.486 & 0.926
& 0.737 & 0.809
& 0.586 & 0.863
& 43904 & 18696 \\

\midrule

DeepFool & Clean
& 0.903 & 0.814
& 0.520 & 0.936
& 0.771 & 0.825
& 0.620 & 0.877
& 44271 & 18328 \\
DeepFool & DeepFool
& 0.905 & 0.815
& 0.521 & 0.937
& 0.774 & 0.825
& 0.622 & 0.877
& 44238 & 18362 \\
DeepFool & FGSM
& 0.917 & 0.822
& 0.532 & 0.947
& 0.810 & 0.825
& 0.642 & 0.881
& 43796 & 18803 \\
DeepFool & PGD
& 0.917 & 0.822
& 0.532 & 0.947
& 0.810 & 0.825
& 0.642 & 0.881
& 43796 & 18803 \\
DeepFool & S\&P
& 0.860 & 0.784
& 0.471 & 0.928
& 0.750 & 0.792
& 0.578 & 0.855
& 42919 & 19680 \\

\midrule

FGSM & Clean
& 0.899 & 0.827
& 0.548 & 0.927
& 0.728 & 0.852
& 0.624 & 0.888
& 46178 & 16421 \\
FGSM & DeepFool
& 0.908 & 0.832
& 0.555 & 0.933
& 0.750 & 0.852
& 0.637 & 0.890
& 45904 & 16696 \\
FGSM & FGSM
& 0.943 & 0.854
& 0.589 & 0.962
& 0.862 & 0.852
& 0.699 & 0.903
& 44526 & 18074 \\
FGSM & PGD
& 0.943 & 0.854
& 0.589 & 0.962
& 0.862 & 0.852
& 0.699 & 0.903
& 44526 & 18074 \\
FGSM & S\&P
& 0.856 & 0.797
& 0.491 & 0.920
& 0.707 & 0.820
& 0.579 & 0.867
& 44813 & 17787 \\

\midrule

PGD & Clean
& 0.899 & 0.825
& 0.541 & 0.929
& 0.735 & 0.847
& 0.623 & 0.886
& 45848 & 16751 \\
PGD & DeepFool
& 0.909 & 0.829
& 0.548 & 0.934
& 0.756 & 0.847
& 0.635 & 0.888
& 45589 & 17011 \\
PGD & FGSM
& 0.945 & 0.852
& 0.583 & 0.965
& 0.873 & 0.847
& 0.699 & 0.902
& 44147 & 18452 \\
PGD & PGD
& 0.945 & 0.852
& 0.583 & 0.965
& 0.873 & 0.847
& 0.699 & 0.902
& 44147 & 18452 \\
PGD & S\&P
& 0.857 & 0.795
& 0.487 & 0.921
& 0.714 & 0.815
& 0.579 & 0.865
& 44504 & 18095 \\

\midrule

S\&P & Clean
& 0.905 & 0.823
& 0.535 & 0.935
& 0.764 & 0.837
& 0.629 & 0.883
& 44990 & 17609 \\
S\&P & DeepFool
& 0.860 & 0.796
& 0.486 & 0.902
& 0.630 & 0.837
& 0.548 & 0.868
& 46640 & 15959 \\
S\&P & FGSM
& 0.864 & 0.802
& 0.498 & 0.909
& 0.658 & 0.837
& 0.567 & 0.872
& 46289 & 16310 \\
S\&P & PGD
& 0.864 & 0.802
& 0.498 & 0.909
& 0.658 & 0.837
& 0.567 & 0.872
& 46289 & 16310 \\
S\&P & S\&P
& 0.865 & 0.793
& 0.483 & 0.927
& 0.742 & 0.805
& 0.585 & 0.862
& 43669 & 18930 \\

\midrule

Mixed & Clean
& 0.901 & 0.822
& 0.534 & 0.932
& 0.750 & 0.839
& 0.624 & 0.883
& 45271 & 17328 \\
Mixed & DeepFool
& 0.907 & 0.824
& 0.537 & 0.934
& 0.760 & 0.839
& 0.629 & 0.884
& 45154 & 17445 \\
Mixed & FGSM
& 0.930 & 0.838
& 0.561 & 0.954
& 0.835 & 0.839
& 0.671 & 0.893
& 44227 & 18372 \\
Mixed & PGD
& 0.930 & 0.838
& 0.561 & 0.954
& 0.835 & 0.839
& 0.671 & 0.893
& 44227 & 18372 \\
Mixed & S\&P
& 0.860 & 0.792
& 0.482 & 0.924
& 0.728 & 0.808
& 0.580 & 0.862
& 43985 & 18615 \\

\bottomrule
\end{tabular}}
\end{table}

\begin{table}[htbp]
\centering
\scriptsize
\setlength{\tabcolsep}{2.5pt}
\caption{Mean performance of the Neural Network model across cross-validation folds under different adversarial training and testing configurations. All values represent fold-averaged metrics and are rounded to three decimal digits.}
\label{tab:nn-mean-adv}
\resizebox{0.85\textwidth}{!}{%
\begin{tabular}{l l c c c c c c c c c c}
\toprule
Training Set & Test Set &
ROC AUC & Accuracy &
Prec$_0$ & Prec$_1$ &
Rec$_0$ & Rec$_1$ &
F1$_0$ & F1$_1$ &
Pred ND & Pred D \\
\midrule

Clean & Clean
& 0.903 & 0.822
& 0.534 & 0.935
& 0.765 & 0.836
& 0.628 & 0.883
& 44911 & 17688 \\
Clean & DeepFool
& 0.873 & 0.798
& 0.490 & 0.906
& 0.644 & 0.836
& 0.555 & 0.869
& 46397 & 16202 \\
Clean & FGSM
& 0.871 & 0.805
& 0.504 & 0.914
& 0.678 & 0.836
& 0.577 & 0.873
& 45982 & 16617 \\
Clean & PGD
& 0.871 & 0.804
& 0.504 & 0.914
& 0.678 & 0.836
& 0.577 & 0.873
& 45984 & 16616 \\
Clean & S\&P
& 0.858 & 0.790
& 0.480 & 0.927
& 0.743 & 0.802
& 0.583 & 0.860
& 43480 & 19120 \\

\midrule

DeepFool & Clean
& 0.901 & 0.825
& 0.542 & 0.933
& 0.753 & 0.843
& 0.629 & 0.885
& 45421 & 17178 \\
DeepFool & DeepFool
& 0.905 & 0.823
& 0.539 & 0.931
& 0.743 & 0.843
& 0.624 & 0.884
& 45537 & 17062 \\
DeepFool & FGSM
& 0.895 & 0.826
& 0.543 & 0.934
& 0.755 & 0.843
& 0.631 & 0.886
& 45387 & 17212 \\
DeepFool & PGD
& 0.896 & 0.826
& 0.543 & 0.934
& 0.756 & 0.843
& 0.631 & 0.886
& 45380 & 17219 \\
DeepFool & S\&P
& 0.856 & 0.795
& 0.487 & 0.925
& 0.730 & 0.810
& 0.583 & 0.864
& 44059 & 18541 \\

\midrule

FGSM & Clean
& 0.899 & 0.840
& 0.579 & 0.925
& 0.711 & 0.872
& 0.637 & 0.898
& 47398 & 15201 \\
FGSM & DeepFool
& 0.891 & 0.830
& 0.560 & 0.913
& 0.659 & 0.872
& 0.605 & 0.892
& 48040 & 14559 \\
FGSM & FGSM
& 0.964 & 0.877
& 0.634 & 0.972
& 0.898 & 0.872
& 0.743 & 0.919
& 45095 & 17504 \\
FGSM & PGD
& 0.965 & 0.877
& 0.634 & 0.973
& 0.899 & 0.872
& 0.743 & 0.919
& 45080 & 17519 \\
FGSM & S\&P
& 0.854 & 0.810
& 0.514 & 0.917
& 0.691 & 0.839
& 0.588 & 0.876
& 45985 & 16614 \\

\midrule

PGD & Clean
& 0.901 & 0.842
& 0.581 & 0.925
& 0.709 & 0.874
& 0.639 & 0.899
& 47537 & 15063 \\
PGD & DeepFool
& 0.902 & 0.838
& 0.574 & 0.920
& 0.689 & 0.874
& 0.626 & 0.896
& 47790 & 14810 \\
PGD & FGSM
& 0.970 & 0.883
& 0.642 & 0.977
& 0.917 & 0.874
& 0.755 & 0.923
& 44976 & 17623 \\
PGD & PGD
& 0.971 & 0.883
& 0.643 & 0.978
& 0.920 & 0.874
& 0.757 & 0.923
& 44940 & 17659 \\
PGD & S\&P
& 0.857 & 0.811
& 0.515 & 0.917
& 0.689 & 0.840
& 0.589 & 0.877
& 46076 & 16524 \\

\midrule

S\&P & Clean
& 0.901 & 0.817
& 0.525 & 0.937
& 0.771 & 0.829
& 0.625 & 0.879
& 44494 & 18105 \\
S\&P & DeepFool
& 0.871 & 0.801
& 0.496 & 0.916
& 0.687 & 0.829
& 0.576 & 0.870
& 45524 & 17076 \\
S\&P & FGSM
& 0.868 & 0.800
& 0.495 & 0.914
& 0.682 & 0.829
& 0.573 & 0.869
& 45586 & 17013 \\
S\&P & PGD
& 0.868 & 0.800
& 0.495 & 0.914
& 0.682 & 0.829
& 0.573 & 0.869
& 45591 & 17009 \\
S\&P & S\&P
& 0.874 & 0.792
& 0.483 & 0.929
& 0.751 & 0.802
& 0.588 & 0.861
& 43413 & 19187 \\

\midrule

Mixed & Clean
& 0.901 & 0.829
& 0.550 & 0.930
& 0.740 & 0.851
& 0.631 & 0.889
& 45990 & 16609 \\
Mixed & DeepFool
& 0.906 & 0.827
& 0.546 & 0.928
& 0.729 & 0.851
& 0.624 & 0.888
& 46127 & 16472 \\
Mixed & FGSM
& 0.943 & 0.852
& 0.585 & 0.960
& 0.855 & 0.851
& 0.694 & 0.902
& 44577 & 18023 \\
Mixed & PGD
& 0.943 & 0.852
& 0.585 & 0.960
& 0.856 & 0.851
& 0.695 & 0.902
& 44558 & 18041 \\
Mixed & S\&P
& 0.868 & 0.800
& 0.495 & 0.923
& 0.719 & 0.820
& 0.586 & 0.868
& 44679 & 17921 \\

\bottomrule
\end{tabular}}
\end{table}

\begin{table}[htbp]
\centering
\scriptsize
\setlength{\tabcolsep}{2.5pt}
\caption{Mean performance of the Transformer model across cross-validation folds under different adversarial training and testing configurations. All values represent fold-averaged metrics and are rounded to three decimal digits.}
\label{tab:ftt-mean-adv}
\resizebox{0.85\textwidth}{!}{%
\begin{tabular}{l l c c c c c c c c c c}
\toprule
Training Set & Test Set &
ROC AUC & Accuracy &
Prec$_0$ & Prec$_1$ &
Rec$_0$ & Rec$_1$ &
F1$_0$ & F1$_1$ &
Pred ND & Pred D \\
\midrule

Clean & Clean
& 0.906 & 0.826
& 0.542 & 0.934
& 0.758 & 0.843
& 0.632 & 0.886
& 45342 & 17258 \\
Clean & DeepFool
& 0.872 & 0.795
& 0.482 & 0.897
& 0.602 & 0.843
& 0.535 & 0.869
& 47261 & 15339 \\
Clean & FGSM
& 0.870 & 0.806
& 0.507 & 0.910
& 0.659 & 0.843
& 0.573 & 0.875
& 46553 & 16046 \\
Clean & PGD
& 0.870 & 0.806
& 0.507 & 0.910
& 0.659 & 0.843
& 0.573 & 0.875
& 46552 & 16047 \\
Clean & S\&P
& 0.865 & 0.796
& 0.488 & 0.927
& 0.739 & 0.810
& 0.588 & 0.864
& 43923 & 18676 \\

\midrule

DeepFool & Clean
& 0.905 & 0.832
& 0.555 & 0.930
& 0.739 & 0.855
& 0.634 & 0.891
& 46188 & 16411 \\
DeepFool & DeepFool
& 0.908 & 0.830
& 0.551 & 0.927
& 0.727 & 0.855
& 0.627 & 0.890
& 46325 & 16274 \\
DeepFool & FGSM
& 0.897 & 0.830
& 0.551 & 0.927
& 0.727 & 0.855
& 0.627 & 0.890
& 46329 & 16271 \\
DeepFool & PGD
& 0.897 & 0.830
& 0.551 & 0.927
& 0.727 & 0.855
& 0.627 & 0.890
& 46329 & 16271 \\
DeepFool & S\&P
& 0.864 & 0.801
& 0.497 & 0.923
& 0.721 & 0.821
& 0.588 & 0.869
& 44706 & 17893 \\

\midrule

FGSM & Clean
& 0.902 & 0.846
& 0.597 & 0.920
& 0.688 & 0.885
& 0.638 & 0.902
& 48337 & 14263 \\
FGSM & DeepFool
& 0.905 & 0.846
& 0.596 & 0.920
& 0.687 & 0.885
& 0.637 & 0.902
& 48347 & 14253 \\
FGSM & FGSM
& 0.975 & 0.893
& 0.665 & 0.980
& 0.927 & 0.885
& 0.774 & 0.930
& 45386 & 17214 \\
FGSM & PGD
& 0.974 & 0.893
& 0.665 & 0.980
& 0.926 & 0.885
& 0.774 & 0.930
& 45400 & 17199 \\
FGSM & S\&P
& 0.860 & 0.815
& 0.526 & 0.914
& 0.671 & 0.851
& 0.589 & 0.881
& 46818 & 15781 \\

\midrule

PGD & Clean
& 0.903 & 0.840
& 0.578 & 0.925
& 0.710 & 0.872
& 0.637 & 0.898
& 47434 & 15165 \\
PGD & DeepFool
& 0.911 & 0.840
& 0.577 & 0.924
& 0.706 & 0.872
& 0.635 & 0.897
& 47478 & 15121 \\
PGD & FGSM
& 0.981 & 0.888
& 0.647 & 0.986
& 0.950 & 0.872
& 0.770 & 0.926
& 44474 & 18125 \\
PGD & PGD
& 0.980 & 0.887
& 0.647 & 0.986
& 0.949 & 0.872
& 0.769 & 0.926
& 44493 & 18107 \\
PGD & S\&P
& 0.860 & 0.810
& 0.513 & 0.918
& 0.692 & 0.838
& 0.589 & 0.876
& 45942 & 16657 \\

\midrule

S\&P & Clean
& 0.904 & 0.827
& 0.544 & 0.932
& 0.750 & 0.845
& 0.630 & 0.887
& 45587 & 17013 \\
S\&P & DeepFool
& 0.865 & 0.795
& 0.482 & 0.895
& 0.592 & 0.845
& 0.530 & 0.869
& 47535 & 15065 \\
S\&P & FGSM
& 0.869 & 0.809
& 0.511 & 0.910
& 0.659 & 0.845
& 0.575 & 0.876
& 46709 & 15890 \\
S\&P & PGD
& 0.869 & 0.809
& 0.511 & 0.910
& 0.659 & 0.845
& 0.575 & 0.876
& 46709 & 15891 \\
S\&P & S\&P
& 0.879 & 0.804
& 0.502 & 0.925
& 0.730 & 0.822
& 0.594 & 0.871
& 44649 & 17950 \\

\midrule

Mixed & Clean
& 0.904 & 0.828
& 0.547 & 0.932
& 0.748 & 0.848
& 0.631 & 0.888
& 45728 & 16871 \\
Mixed & DeepFool
& 0.913 & 0.827
& 0.544 & 0.930
& 0.740 & 0.848
& 0.627 & 0.887
& 45818 & 16781 \\
Mixed & FGSM
& 0.974 & 0.866
& 0.602 & 0.983
& 0.939 & 0.848
& 0.734 & 0.910
& 43366 & 19233 \\
Mixed & PGD
& 0.974 & 0.865
& 0.602 & 0.982
& 0.938 & 0.848
& 0.733 & 0.910
& 43381 & 19218 \\
Mixed & S\&P
& 0.876 & 0.802
& 0.499 & 0.925
& 0.728 & 0.820
& 0.592 & 0.869
& 44586 & 18014 \\

\bottomrule
\end{tabular}}
\end{table}

\clearpage

\section{External replication on a second P2P dataset (Prosper)}
\label{appendixB}

To assess whether the robustness and cross-attack transfer patterns reported in
Section~\ref{sec:results} are specific to Lending Club or reflect more general
behaviour of adversarially trained tabular credit models, we replicate the entire
study on an independent P2P dataset from the Prosper marketplace.\footnote{Prosper
Marketplace, \url{https://www.prosper.com/}.} The replication is deliberately
constrained to be method-identical: the same three model families
(logistic regression, feed-forward neural network, FT-Transformer), the same four
attacks (FGSM, PGD, DeepFool, S\&P) plus the mixed regime, the same
$80/20$ split with stratified $5$-fold cross-validation, the same
$\varepsilon=\text{amount}=0.1$ budgets, and the same class-weighted training. Only
the feature space is re-mapped to the Prosper schema. The mutable/immutable
partition follows the identical philosophy used for Lending Club: the four
self-reported financial variables (\texttt{annual\_inc}, \texttt{dti},
\texttt{revol\_bal}, \texttt{revol\_util}) and two self-reported categorical blocks
(\texttt{purpose}, \texttt{addr\_state} in place of \texttt{zip\_code}) are
perturbable, while platform-generated scores (\texttt{ProsperRating},
\texttt{ProsperScore}), immutable credit-history facts, and contractually fixed
terms are blocked via gradient masking and structurally valid one-hot swaps. The
full feature-level constraint table is available in the project repository.

Two structural differences from the main analysis should be kept in mind when
reading the results. First, the Prosper subset is roughly an order of magnitude
smaller (test folds of $\approx 3{,}900$ observations versus $\approx 62{,}600$ for
Lending Club), so fold-to-fold variability is larger. Second, and more importantly,
the clean discriminative signal is weaker: all three models attain clean ROC~AUC in
the $0.70$--$0.71$ range (Table~\ref{tab:prosper-baseline}), well below the
$0.90$--$0.91$ obtained on Lending Club (Table~\ref{tab:model-comparison}). The
replication therefore tests whether the \emph{qualitative} robustness structure
survives on a harder, differently constructed dataset, rather than whether absolute
performance is reproduced.

\begin{table}[htbp]
  \centering
  \scriptsize
  \setlength{\tabcolsep}{3pt}
  \caption{Clean-train/clean-test performance of the three model families on the
  Prosper dataset (fold-averaged means). Compare with Table~\ref{tab:model-comparison}.}
  \label{tab:prosper-baseline}
  \begin{tabular}{l *{10}{c}}
    \toprule
    Model
      & ROC AUC & Accuracy
      & Prec$_0$ & Rec$_0$ & F1$_0$
      & Prec$_1$ & Rec$_1$ & F1$_1$
      & Pred.\ ND & Pred.\ D \\
    \midrule
    Logistic Regression
      & 0.702 & 0.633
      & 0.379 & 0.674 & 0.485
      & 0.847 & 0.619 & 0.715
      & 2{,}113.6 & 1{,}769.6 \\
    Neural Network
      & 0.698 & 0.641
      & 0.381 & 0.646 & 0.479
      & 0.840 & 0.639 & 0.726
      & 2{,}198.6 & 1{,}684.6 \\
    FT-Transformer
      & 0.711 & 0.691
      & 0.423 & 0.546 & 0.474
      & 0.827 & 0.740 & 0.780
      & 2{,}590.0 & 1{,}293.0 \\
    \bottomrule
  \end{tabular}
\end{table}

Table~\ref{tab:prosper-roc-grid} reports the complete train$\times$test grid using
ROC~AUC, the primary ranking metric in the main text, as fold-averaged mean with the
cross-fold standard deviation in parentheses. Matched attack--defence cells (the
diagonal of the attack rows) are shown in bold.

\begin{table}[htbp]
\centering
\scriptsize
\setlength{\tabcolsep}{4pt}
\caption{Cross-attack robustness grid on the Prosper dataset: ROC~AUC (fold-averaged
mean, cross-fold standard deviation in parentheses) for every training regime (rows)
$\times$ test perturbation (columns), for the three model families. Bold entries mark
matched attack--defence cells. This is the Prosper analogue of
Tables~\ref{tab:logit-adv-grid}--\ref{tab:transformer-adv-cv}.}
\label{tab:prosper-roc-grid}
\resizebox{\textwidth}{!}{%
\begin{tabular}{l c c c c c}
\toprule
Training Set & Clean & DeepFool & FGSM & PGD & S\&P \\
\midrule
\multicolumn{6}{l}{\textit{Logistic Regression}} \\
\midrule
Clean     & 0.702\,(0.006) & 0.635\,(0.036) & 0.666\,(0.020) & 0.666\,(0.020) & 0.700\,(0.007) \\
DeepFool  & 0.685\,(0.019) & \textbf{0.724\,(0.006)} & 0.712\,(0.008) & 0.712\,(0.008) & 0.684\,(0.019) \\
FGSM      & 0.695\,(0.012) & 0.706\,(0.013) & \textbf{0.701\,(0.011)} & 0.701\,(0.011) & 0.693\,(0.012) \\
PGD       & 0.694\,(0.010) & 0.693\,(0.025) & 0.695\,(0.019) & \textbf{0.695\,(0.019)} & 0.692\,(0.010) \\
S\&P      & 0.702\,(0.011) & 0.620\,(0.028) & 0.660\,(0.011) & 0.660\,(0.011) & \textbf{0.701\,(0.011)} \\
Mixed     & 0.689\,(0.013) & 0.693\,(0.012) & 0.692\,(0.011) & 0.692\,(0.011) & 0.687\,(0.013) \\
\midrule
\multicolumn{6}{l}{\textit{Neural Network}} \\
\midrule
Clean     & 0.698\,(0.023) & 0.545\,(0.061) & 0.588\,(0.029) & 0.587\,(0.030) & 0.690\,(0.020) \\
DeepFool  & 0.697\,(0.010) & \textbf{0.756\,(0.010)} & 0.728\,(0.020) & 0.726\,(0.023) & 0.690\,(0.011) \\
FGSM      & 0.705\,(0.008) & 0.767\,(0.005) & \textbf{0.811\,(0.014)} & 0.806\,(0.020) & 0.697\,(0.008) \\
PGD       & 0.703\,(0.010) & 0.766\,(0.007) & 0.827\,(0.020) & \textbf{0.822\,(0.027)} & 0.695\,(0.010) \\
S\&P      & 0.709\,(0.008) & 0.539\,(0.036) & 0.589\,(0.018) & 0.588\,(0.018) & \textbf{0.701\,(0.007)} \\
Mixed     & 0.699\,(0.010) & 0.759\,(0.009) & 0.761\,(0.015) & 0.758\,(0.020) & 0.691\,(0.008) \\
\midrule
\multicolumn{6}{l}{\textit{FT-Transformer}} \\
\midrule
Clean     & 0.711\,(0.010) & 0.638\,(0.035) & 0.643\,(0.015) & 0.643\,(0.015) & 0.708\,(0.011) \\
DeepFool  & 0.705\,(0.010) & \textbf{0.712\,(0.026)} & 0.708\,(0.029) & 0.708\,(0.029) & 0.703\,(0.011) \\
FGSM      & 0.699\,(0.010) & 0.706\,(0.013) & \textbf{0.712\,(0.014)} & 0.712\,(0.015) & 0.696\,(0.011) \\
PGD       & 0.699\,(0.011) & 0.707\,(0.013) & 0.715\,(0.015) & \textbf{0.716\,(0.016)} & 0.696\,(0.012) \\
S\&P      & 0.709\,(0.012) & 0.617\,(0.062) & 0.632\,(0.020) & 0.632\,(0.021) & \textbf{0.706\,(0.013)} \\
Mixed     & 0.701\,(0.013) & 0.699\,(0.014) & 0.704\,(0.013) & 0.704\,(0.013) & 0.698\,(0.015) \\
\bottomrule
\end{tabular}}
\end{table}

\paragraph{What replicates} Four of the main findings reappear cleanly. First,
gradient-based attacks are the most damaging to clean-trained models, whereas S\&P is
comparatively mild: the clean-trained neural network falls from ROC~AUC $0.698$ to
$0.588$/$0.587$ under FGSM/PGD and to $0.545$ under DeepFool, but only to $0.690$
under S\&P (Table~\ref{tab:prosper-roc-grid}), echoing the main-text ordering in which
S\&P produced a broad but shallow decline. Second, matched-attack adversarial
training recovers and improves discrimination, and does so most strongly for the
gradient family and the higher-capacity neural network: FGSM- and PGD-trained neural
networks reach ROC~AUC $0.811$ and $0.822$ on their matched tests, far above the
clean-trained $0.588$ baseline, mirroring the large matched-training gains reported
for Lending Club. Third, FGSM and PGD remain mutually interchangeable, with near-identical
columns and cross-cells (e.g., FGSM-train/PGD-test $0.806$ vs.\ PGD-train/FGSM-test
$0.827$ for the neural network), confirming that their shared gradient structure, and
not dataset-specific artefacts, drives the transfer. Fourth, cross-family transfer is
weak in both directions: S\&P training leaves gradient robustness at clean-model
levels ($0.589$/$0.588$ for the neural network), and gradient training barely moves
S\&P performance, exactly as in the main analysis.

\paragraph{What differs, and why it strengthens the generalisation claim} Two
departures are worth stating plainly. The mixed regime is again the most balanced
policy, delivering solid performance across all five test conditions
(e.g., the neural network at $0.699$--$0.761$ with no catastrophic cell) while
preserving clean accuracy, but on this weaker-signal dataset it no longer strictly
dominates the best single-attack model, since matched gradient training attains
higher gradient-test scores. Likewise, the architecture that benefits most
shifts: on Prosper the feed-forward network shows the largest adversarial-training
gains, whereas the FT-Transformer, which led on Lending Club, improves only modestly
(matched cells $0.712$--$0.716$), and logistic regression again benefits least, in
line with its main-text behaviour. The absolute levels and the best-performing architecture change, but the ordering of the training regimes does not: matched-attack training remains the strongest defence, transfer stays high within the gradient family and low across families, and mixed training remains the most balanced choice. 
\end{document}